\documentclass[twocolumn]{svjour3}          
\smartqed  
\usepackage[pdftex]{graphicx}
\usepackage{subfig}

\usepackage{natbib}
\usepackage{multicol}
\usepackage{amsfonts}
\usepackage{mathrsfs}

\usepackage{fancyhdr} 
 \journalname{myjournal}

\usepackage{algorithmic}

\makeatletter
\newcounter{algorithmbis}
\renewcommand{\thealgorithmbis}{\thesection.\arabic{algorithmbis}}
\def\algorithmbis{\@ifnextchar[{\@algorithmbisa}{\@algorithmbisb}}
\def\@algorithmbisa[#1]{%
  \refstepcounter{algorithmbis}
  \trivlist
  \leftmargin\z@
  \itemindent\z@
  \labelsep\z@
  \item[\parbox{\linewidth}{%
    \hrule
    \hrule
    \noindent\strut\textbf{Algorithm \thealgorithmbis} #1
    \hrule
  }]\hfil\vskip0em%
}
\def\@algorithmbisb{\@algorithmbisa[]}

\makeatother

\newcommand{\captionfonts}{\footnotesize}

\makeatletter  
\long\def\@makecaption#1#2{%
  \vskip\abovecaptionskip
  \sbox\@tempboxa{{\captionfonts #1: #2}}%
  \ifdim \wd\@tempboxa >\hsize
    {\captionfonts #1: #2\par}
  \else
    \hbox to\hsize{\hfil\box\@tempboxa\hfil}%
  \fi
  \vskip\belowcaptionskip}
\makeatother   

\usepackage[pdftex]{hyperref}

\begin{document}

\title{Socially Guided Intrinsic Motivation \\
for Robot Learning of Motor Skills}


\author{Sao~Mai~Nguyen         \and
        Pierre-Yves~Oudeyer 
}


\institute{Flowers Team, INRIA and ENSTA ParisTech, France}

\date{Received: 2012 / Accepted: 2013}

\maketitle
\thispagestyle{empty}
\thispagestyle{fancy}
\lhead{}
\lfoot{}
\cfoot{
\texttt{\scriptsize{S. M. Nguyen and P.-Y. Oudeyer. Socially Guided Intrinsic Motivation for Robot Learning of Motor Skills, Autonomous Robots, 2014,  doi:http://dx.doi.org/10.1007/s10514-013-9339-y,  36(3):273-294 . }}
\vspace{0pt}}
\rfoot{}

\begin{abstract}
This paper presents a technical approach to robot learning of motor skills which combines active intrinsically motivated learning with imitation learning. Our architecture, called \textbf{SGIM-D}, allows efficient learning of high-dimensional continuous sensorimotor inverse models in robots, and in particular learns distributions of parameterised motor policies that solve a corresponding distribution of parameterised goals/tasks. This is made possible by the technical integration of imitation learning techniques within an algorithm for learning inverse models that relies on active goal babbling. After reviewing social learning and intrinsic motivation approaches to action learning, we describe the general framework of our algorithm, before detailing its architecture. In an experiment where a robot arm has to learn to use a flexible fishing line , we illustrate that SGIM-D efficiently combines the advantages of social learning and intrinsic motivation and benefits from human demonstration properties to learn how to produce varied outcomes in the environment, while developing more precise control policies in large spaces. 

\end{abstract}

\begin{figure}
\centering
\includegraphics[width=0.5\textwidth]{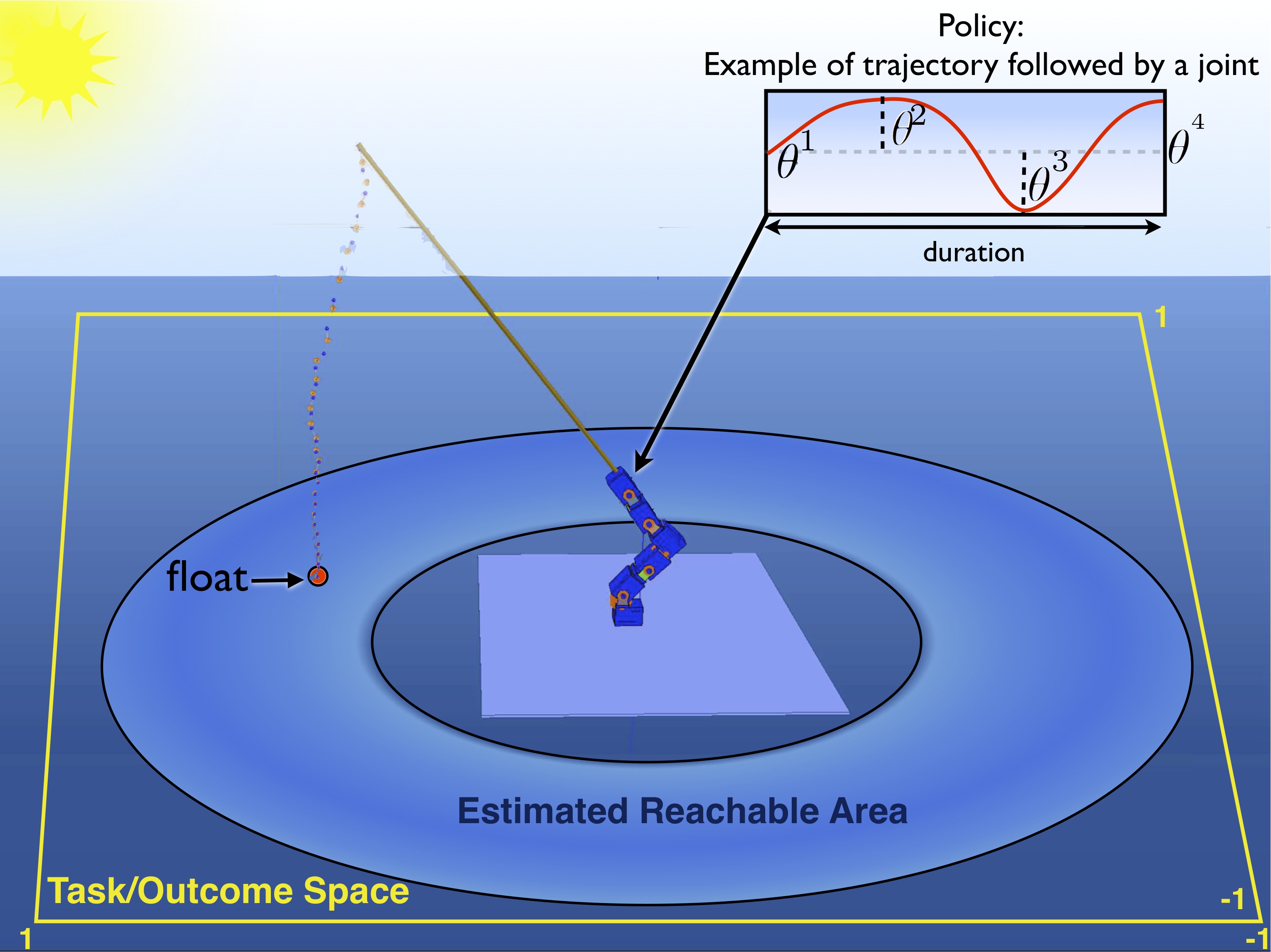}
\caption{Experimental setup with a robot arm holding a fishing can with a flexible wire (simulated by 30 free revolute joints). The robot can produce a movement of its 6 DOF arm by setting the real number values $\theta$ of its 25 dimensional motor primitive. Then, it can observe the effect/outcome of such a movement, by observing where the float has arrived in the task space, i.e. on the surface of the water which is a 2D space. Using SGIM-D, which combines intrinsically motivated active learning and human demonstration, the robot has to learn the complex inverse model mapping all goals/outcomes to the adequate 25 dimensional parameters of motor movement.}
\label{FishingRod}
\vspace{-0.6cm}
\end{figure}

In this article, we first review approaches to life-long skill learning in section 1, set the framework of our approach in section 2, then describe our algorithm in section 3, before implementing it for a fishing robot (fig. \ref{FishingRod}).

\section{Approaches to Skill Learning in Adaptive Personal Robots}

The promise of personal robots operating in human environments to interact with people on a daily basis points out the importance of adaptivity. The robot can no longer simply 
reproduce actions predesigned in factories. It needs to adapt to its changing and open-ended environment, match its behaviour and learn new skills as the environment and users' needs change.

Yet, robot learning of new action skills is a difficult problem because their sensorimotor spaces are large and high-dimensional, and at the same time their physical embedding allows only limited time for collecting training data. For example, random motor exploration is bound to fail for building forward or inverse models through regression in high-dimension \citep{Baranes2013RAS}. Thus, learning must be associated to mechanisms for guided exploration. Exploration strategies  developed in the recent years can be classified into two broad interacting families: 1) socially guided exploration \citep{Nehaniv04,Billard2007RobotProgrammingby,Argall2009RAS}; 2) internally guided exploration and in particular intrinsically motivated exploration \citep{Schmidhuber91,Barto04,Oudeyer07}.

\subsection{Socially Guided Exploration and Imitation Learning}
In order to build a robot that can learn and adapt to human environment, the most straightforward way might be the knowledge transfer  from a human into a machine.  Behavioural psychology studies   \citep{Whiten2000CS,Tomasello2007DS} highlight the processes through which the behaviour of an individual B may come to be like A's, such as mimicry, stimulus enhancement, imitation or emulation.

Learning a policy from demonstrations provided by a teacher is commonly referred to as Programming by Demonstration (PbD) or imitation learning \citep{Nehaniv04,Billard2007RobotProgrammingby,Argall2009RAS}.
PbD targets an implicit means of training a machine, such that explicit and tedious programming of a task by a human user can be minimised. It is an intuitive medium of communication for humans, who already use demonstrations to teach other humans. It can in principle offer a natural means of teaching machines that would be accessible to non experts. For instance trajectory and keyframe demonstrations have been shown to be efficient and easy to use for non-experts \citep{Akgun2012ICHI}.

That is why several  works  incorporate human input to guide the robot learning process, such as in some examples of trajectory-based PbD where statistical regression techniques are used to model the invariances of demonstrated movements \citep{Billard2007RobotProgrammingby,GrollmanJenkins2008icra,Chernova2009JAIR, Lopes2009AB, Cederborg10, PbDCalinon,Calinon07,Peters2008NN}, or inverse reinforcement learning approaches \citep{Abbeel04icml, Verma06ImitInfer,mangin12hbu} where one attempts to achieve goal imitation by inferring the hidden cost function maximised by the demonstrated movement \citep{lopes10imitationchapter}. Prior works have  also given a human trainer control of  the reinforcement learning reward \citep{Blumberg:2002:ILI:566654.566597,Robotic-clicker-training}, provide advice \citep{A-teaching-method-for-reinforcement-leClouse-J.-and-Utgoff-P.-arning}, or  teleoperate the agent during training \citep{Effective-reinforcement-learning-for-mobile-robots}.

In these systems, learning has been strongly relying on the involvement of the human user. However, the more dependent on the human the system, the more challenging learning from interactions with a human is, due to limitations such as human patience, teaching dataset sparsity, the absence of teaching for some subspaces, ambiguous and suboptimal human input, correspondence problems, etc, as highlighted in \citep{Nehaniv2007}. This is one of the reasons why in most approaches to robot learning of motor skills, either in trajectory based approaches or inverse reinforcement learning, only a few movements or motor policies where learnt in any single studies.

Increasing the learner's autonomy from human guidance could address these limitations. This is the case of approaches based on more autonomous learning techniques, such as intrinsically motivated exploration methods.
 
 \subsection{Intrinsically Motivated Exploration and Active Learning}

Approaches to robot skill learning based on optimisation and reinforcement learning techniques have been widely studied recently, where one has assumed that an engineer provides manually a reward function that is associated to a pre-defined specific task \citep{kober2011policy,Peters2008NN,Schaal2003PTRSLSBS,Stulp2011H}. Once the reward function is defined, techniques allowing efficient and fast use of training data have been elaborated, such as natural-actor critic architectures \citep{Peters2008NN}, path integral approaches \citep{Theodorou_RAIIC_2010} or advanced Black Box optimisation techniques \citep{stulp:hal13}. While these techniques may seem to rely less on the human expert, they still require an engineer to provide a specific reward function associated to each new particular task to learn. 

In order to allow robots to learn more autonomously a wider diversity of tasks, defined here as goals in a parameterised task space, methods have been devised for learning forward and inverse models, relating a space of parameterised motor policies with a space of parameterised tasks. Once learnt, these forward and inverse models can then be used in conjunction with for example planning methods in order to reach goals. Yet, exploration is a fundamental challenge to achieve the autonomous learning of such forward and inverse models in high-dimensional robots. This is why methods of active exploration and learning have recently been developed in the fields of developmental robotics and robot learning \citep{Oudeyer10b}, reusing some of the concepts elaborated in the statistical active learning framework \citep{Fedorov72,Cohn96,Roy01}. These methods are inspired by \textit{intrinsic motivation} in psychology \citep{Deci85} which trigger spontaneous exploration and curiosity in humans. A first family of such active learning methods, shown to be efficient in spaces up to around 15 continuous dimensions in robots, is called knowledge-based approaches \citep{Oudeyer2007FN,Baldassare11}: parameters of motor policies are chosen for experimentation so that the observed consequences in the task space provide maximal improvement of the quality of the learned forward model, which is then inverted for control when needed \citep{Oudeyer07,Schmidhuber2010ITAMD,Schmidhuber91}. Yet, these methods were shown to become inefficient when dimension increases \citep{Baranes2013RAS}, and these limitations where addressed by competence-based approaches where instead of performing active motor babbling, parameterised tasks were actively sampled through active goal babbling, then generating lower-level goal directed exploration. Goal babbling has been shown recently to considerably fasten learning by exploiting the sensorimotor redundancies and the lower dimensionality of task spaces \citep{Baranes2013RAS, Rolf2010ITAMD,Baranes2010IICIRS}. For example, with the SAGG-RIAC architecture we re-use in this article, it was shown how robots could learn omnidirectional quadruped walking (thus learning to find the parameters of motor policies to achieve the whole variety of possible displacement tasks) or learn inverse arm kinematics with several dozen dimensions (thus learning the parameters of motor policies to reach all spatial goals possible in the visual task space) \citep{Baranes2013RAS}.

However, such active exploration methods for learning forward and inverse models still have limitations. In particular, they address only partially the challenges of unlearnability and unboundedness \citep{OudeyerIM13}, which rises with the use of real high-dimensional bodies with continuous sensorimotor channels and an open-ended environment. Indeed, computing meaningful measures of interest first requires a sampling density which decreases its efficiency as dimensionality grows (curse of dimensionality\citep{Bishop07}). Without additional mechanisms, the identification of learnable zones with knowledge or competence progress becomes less and less efficient as dimensionality grows. The second limitation relates to unboundedness:  whatever the measure of interest used, if it is only based on the evaluation of performances of predictive models or of skills, exploring inside all localities in a lifetime is impossible. Therefore, complementary developmental mechanisms need to constrain the growth of the size and complexity of practically explorable spaces, by introducing self-limits in the unbounded world and/or drive them rapidly toward learnable subspaces \citep{OudeyerIM13}. We argue that social guidance, leveraging knowledge and skills of others, can be key for bootstrapping the intrinsically motivated learning of such models. For example, adequate human demonstration of skills, as we will show in this article, can help the learner to identify which part of the task space are reachable and learnable, as well as to provide examples of motor trajectories useful to reach particular goals, and which can be further explored by the robot to reach self-determined nearby goals.

\subsection{Combining Intrinsically Motivated and Socially Guided Exploration}

Thus, while intrinsic motivation and socially guided learning have so far often been studied separately in developmental robotics and robot learning literature, we believe their integration has high potential.
Their combination could push the respective limits of each family of exploration mechanisms we stated above.

Social guidance can drive a learner into new intrinsically motivating spaces or activities which it may continue to explore alone and for their own sake, but might have discovered only due to social guidance. For example, in the experiment we will present, random uniform exploration of the space of movements has low probability to reach certain areas with the float. Yet, a human may demonstrate early on to the robot specific movements that allow to reach such areas, and then the robot may later on explore variations of these movements through curiosity, allowing the reaching of goals close to these areas.

Conversely, intrinsically motivated learning can build on information provided by human demonstrators/teachers, such as examples of movements or goals to reach, to then spontaneously explore novel movements allowing to reach similar goals in a refined manner or to reach other self-defined goals with the help of these bootstrapping structure provided by humans. In principle, as human demonstration are only used as a bias for further autonomous exploration, intrinsically motivated learning can even use information from human teachers with limited skills, and improve over these demonstrated skills by learning to achieve a higher diversity of goals with more efficient movements.

Thus, while self-exploration alone tends to result in a broader repertoire of skills (i.e. capacity to reach many goals in a task space), and while exploration guided by a human teacher tends to be more specialised and resulting in fewer tasks that are learnt faster, combining both can bring out a system that acquires a diversity of skills with fast bootstrapping thanks to human guidance, and the possibility on the long-term to bias the system towards learning more precisely skills in the preferred areas of the user.

The combination of autonomous learning and imitation learning of continuous high-dimensional motor skills was previously studied in \citep{kober2011policy,Peters2008NN,Schaal2003PTRSLSBS,Stulp2011H}, but this was done only in the context of reinforcement learning one skill, defined as one goal in a task space, and did not rely on active intrinsically motivated learning of forward or inverse models. For example,  \citep{kober2011policy} presented algorithms that allow a robot to learn how to throw a ball at a pre-specified location, by finding adequate parameters of a motor primitive using a human demonstration as bootstrapping and then further optimisation through episodic reinforcement learning. Recently, extensions of these approaches have been presented to allow a robot to generalise motor primitives to novel goals that are close to a set of goals previously learnt with these methods, and leveraging regression techniques \citep{Kober2012AR, Silva20122ICMLI2}. For example, in \citep{Kober2012AR}, a robot can generalise to throw a ball close to a few goals it has already learnt. Yet, in these works, a human engineer has to provide manually a repertoire of goals, and the robot is not able to learn parameters of motor primitives to reach goals that are far away these pre-specified goals. Also, no method for active learning were used in \citep{Kober2012AR, Silva20122ICMLI2}.

A combination of social learning with intrinsic motivational drives was proposed and studied by Thomaz et al. \citep{Thomaz2008CS,Thomaz2006}, with a system called Socially Guided Exploration. In this work, a robot was capable to learn several skills defined as sequences of discrete actions, and as a result of both social dialogue with a human and self-exploration using a hierarchical reinforcement learning algorithm. The focus of this study was on the qualitative dynamics of learning and teaching in the flow of human-robot interaction, and on the design of a full integrated cognitive architecture. While a physical robot was used, the state of the environment as well as robot actions were discrete and few in number. Also, since it was not the focus of these studies, the mechanisms for active learning, for e.g. measuring novelty and mastery, were kept rudimentary and tailored for small discrete state-action spaces. 

In this article, we will present a system, called  Socially Guided Intrinsic Motivation by Demonstration (\textbf{SGIM-D}), that allows a robot to learn a diverse repertoire of parameterised motor primitives, in high-dimensional continuous spaces similar to those used in \citep{kober2011policy,Peters2008NN,Schaal2003PTRSLSBS,Stulp2011H,Kober2012AR, Silva20122ICMLI2}, but allowing to reach a diversity of goals which spans the whole reachable task space. This system will re-use regression techniques allowing to generalise motor primitives to goals close to previously learnt goals, like in \citep{Kober2012AR, Silva20122ICMLI2}, but will allow to self-generate and learn actively goals that are also far from those given by humans. This system will also leverage efficient techniques for active learning of inverse models using goal babbling \citep{Baranes2013RAS, Rolf2010ITAMD,Baranes2010IICIRS}, but extend them with a technical integration with robot learning by demonstration techniques \citep{Billard2007RobotProgrammingby}. Thus, while the combination of social guidance and intrinsic motivation is similar in spirit to the one explored in \citep{Thomaz2008CS}, it will be technically very different and applied to learning sensorimotor skills in continuous high-dimensional spaces more alike the work in \citep{Kober2012AR, Silva20122ICMLI2,Stulp2011H}. 

The next section describes the general framework of SGIM-D. Section 3 details the design of the algorithm. Then, we present our application experiment, the methods used to evaluate our algorithm and finally the experimental results.

\section{General Approach for Socially Guided Intrinsic Motivation}

To better integrate programming by demonstration and intrinsic motivation, we need first to formalise our problem. We then present an overview of our SGIM-D algorithm, before motivating our choice with the statement of our requirements by a detailed examination of different types of social interaction, and  the intrinsic motivation algorithm that we use.

\subsection{Problem Statement and Assumptions}

\label{formalisation}
Csibra'’s theory of human action serves as inspiration for our problem statement. A series of experiments finds that infants connect actions not only to their antecedents but also to their consequents \citep{Csibra2003PTRSLSBS,Csibra2007AP}. 
 Thus, every learning episode can be described as [context][policy][outcome].
We place ourselves in an episodic motor learning framework \citep{Kober2012AR, Silva20122ICMLI2,Stulp2011H}, where a robot is provided with a parameterised encoding of a task space (i.e. it perceives the effect of its movement as a vector or real numbers, e.g. where the ball arrives) as well as a parameterised encoding of movement (i.e. a movement is specified by a vector of real numbers which are parameters of a constrained lower-level motor controller, also called motor primitive). Motor primitives consist in this study in innate or acquired neurally embedded motor and muscle synergies used by humans for control \citep{dAvella2006JN,Weiss2004JN}.
The robot has to learn the inverse model mapping all goals in the task space to corresponding adequate parameters of movements. High-dimensionality in this setting concerns the dimensionality of the vector of parameters for producing movements, which can be different from the actual number of degrees of freedom of the robot since motor primitives control the time evolution of values in each degree of freedom, and this time evolution can be encoded with multiple parameters. For example, in the fishing experiment below, a robot produces a movement of its 6 DOF arm by setting the real number values of its 25 dimensional motor primitives, which controls the evolution of DOFs values by settings targets at different times (global duration being also one of these parameters). Then, it can observe the outcome of such a movement by observing where the float has arrived in the task space, i.e. on the surface of the water which is a 2D value. Using SGIM-D, and thus combining intrinsically motivated learning and human demonstration, the robot has to learn the complex inverse model mapping all goals/tasks (i.e. 2D targets on the water) to adequate parameters of motor movement.
 
 \begin{figure}
  \centering
  \includegraphics[width=0.5\textwidth]{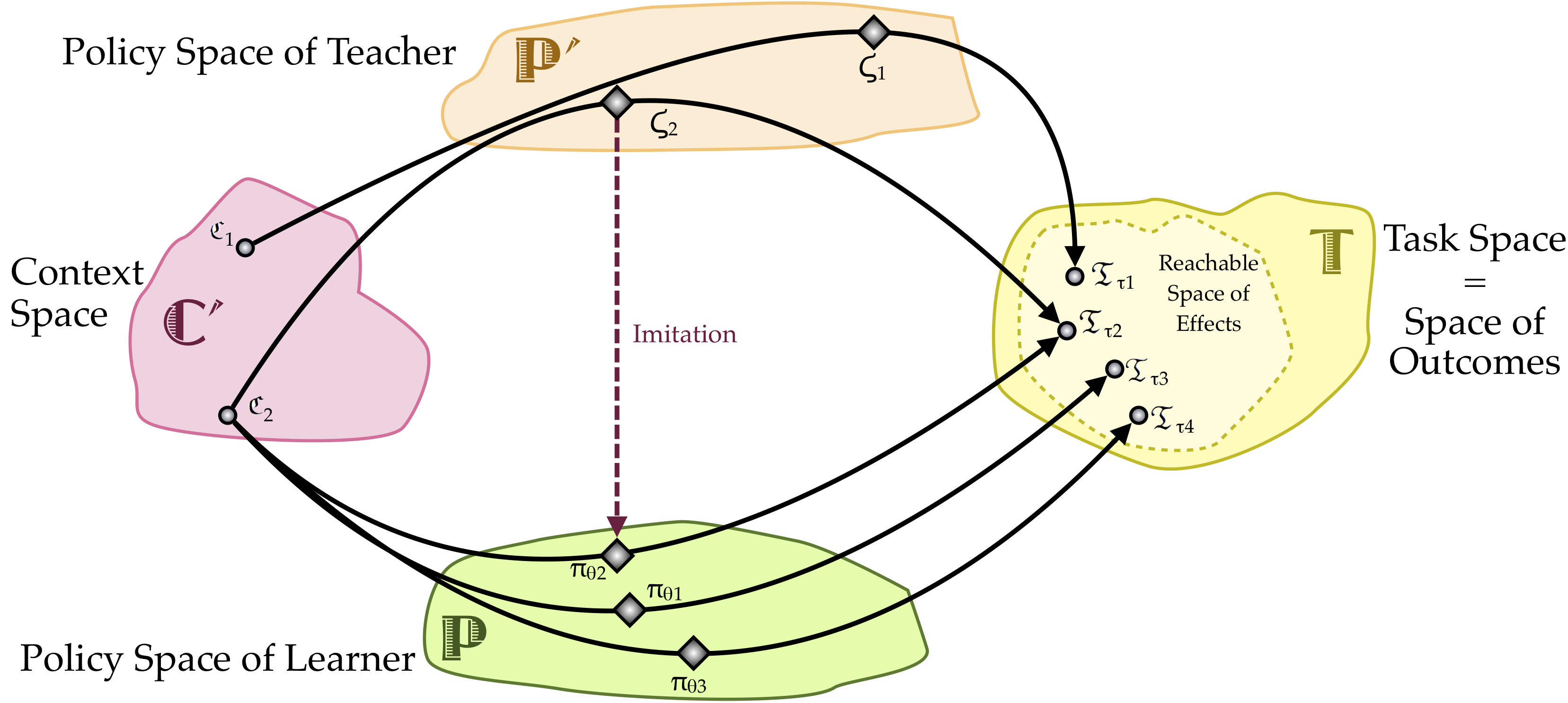}

\caption{Representation of the problem. The environment can evolve from context state $\mathfrak{C}$ to an outcome state $\mathfrak{T}$ by means of the learner's actions with policy $\pi$ or the teacher's $\zeta$. The learner and the teacher have a priori different policy spaces.  The learner estimates $L^{-1}:  \mathfrak{C} \times \mathfrak{T} \mapsto \pi $. By imitation, the learner can take advantage of the demonstrations  (c,$\zeta$,$\tau_d$) of the teacher to improve its estimation $L^{-1}$.}
\label{DataFlow}
\end{figure}

More formally, let us consider an agent learning motor skills, i.e. how to induce any possible goal/task/outcome $\mathfrak{T} \in \mathbb{T}$ from given contexts states $\mathfrak{C} \in \mathbb{C}$ with motor programs $\pi  \in \mathbb{P}$.
  We parameterise the context space with parameters  $c \in C$, and the task space with parameters $\tau \in T$. We define a distance measure $J$ on $T \times T$.
A policy $\pi_\theta$  is described by motor primitives parameterised by $\theta \in \Pi$. From a context $c \in C$, the outcome of policy $\pi_\theta$ is $\tau = M(c, \theta)$, where  the mapping $M: C \times \Pi  \to T$ describes the environment.
 The association $(c, \theta, \tau)$ corresponds to a learning exemplar that will be memorised.

    The performance of a policy $\pi_\theta$ at completing a goal/task $\tau$ from context $c$ is measured by the distance $J(\tau, M(c, \theta))$ between $\tau$ and the outcome of $\pi_\theta$.
The agent focuses on learning the inverse model and builds its estimate $\mathbf{L^{-1}}: C \times T \to \Pi $.  We note that the inverse of the model, $M^{-1}: C \times T \to \Pi$ might not be a function, for $M$ can be redundant. Though, our learner builds a function $\mathbf{L^{-1}}$ that finds at least one adequate policy to complete every goal/task $\tau$ from contexts $c$.  
In sum, it endeavours to minimise :
  \begin {eqnarray}
I =   \int_{\tau \in T, c \in C} P(\tau)J(\tau,M(c, L^{-1}(c,\tau))) d\tau dc
  \end{eqnarray}
where $P(\tau)$ is a probability density distribution over $T$. A priori unknown to the learner, $P(\tau)$ can describe the probability of $\tau$ occurring or the reachable space or a region of interest.
  
  Note that we have described our method without specifying a particular choice of policy representation, learning algorithm or task space properties. These designs can indeed be decided according to the application at hand.

Globally, the learner tries to learn to reach all reachable goals/outcomes $\tau$, and to generalise on the whole task space. 

This problem statement enables a description of an active learning algorithm merging intrinsic motivation with social learning with teacher's demonstrations. We thus design the Socially Guided Intrinsic Motivation by Demonstration (\textbf{SGIM-D})  algorithm which alternates between two strategies.

 \subsection{SGIM-D Overview}
\begin{figure*}
  \centering
  \includegraphics[width=\textwidth]{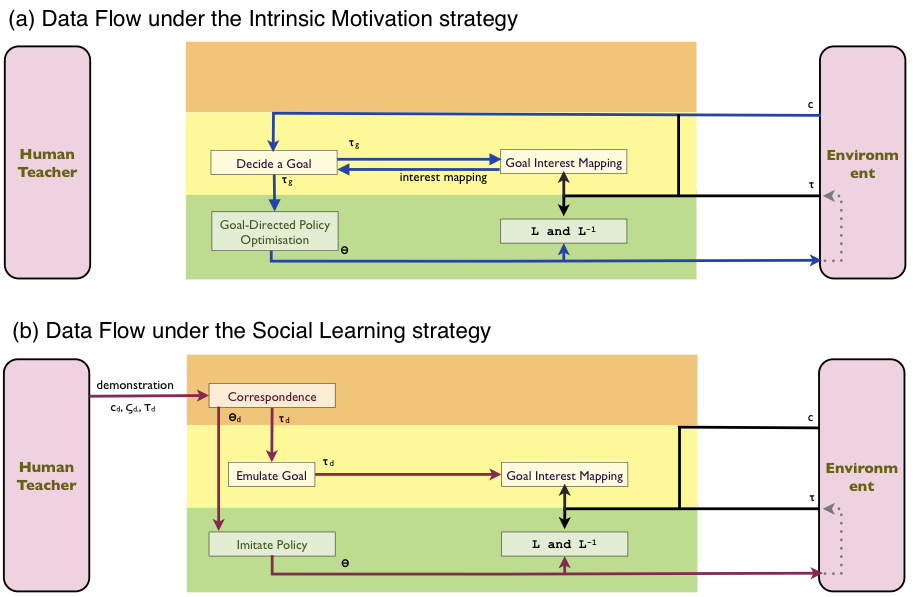}

\caption{Data flow of the SGIM-D learner with its environment and teacher. (a):  intrinsic motivation strategy. (b): social learning strategy. For the details of these graphs, please refer to section 3.}
\label{DataFlow}
\end{figure*} 

SGIM-D improves its estimation $\mathbf{L^{-1}}$ to maximise $I$ both by self-exploring the policy and task space and by imitating  demonstrations (c,$\zeta$,$\tau_d$).

  For the intrinsic motivation strategy (fig. \ref{DataFlow}a),  a wide variety of  intrinsic motivation algorithms have been  developed based on different formal measures of interestingness: minimisation of the prediction error, local density of already sampled points, decrease of the global variance, minimisation of the model uncertainty...\citep{Barto2004IICDL,Oudeyer2011}.
One of the state-of-the-art algorithms, the Self-Adaptive Goal Generation-Robust Intelligent Adaptive Curiosity (SAGG-RIAC), is an implementation of intrinsic motivations based on measures of competence progress \citep{Baranes2013RAS}. It efficiently learns forward and inverse models to reach a wide range of goals in continuous high-dimensional spaces including both easy and unlearnable subparts (see \citep{Rolf2010ITAMD} for another related goal exploration algorithm). 
Moreover, its hierarchical structure proposes 2 levels of learning targeting the task and policy spaces respectively. 
Its goal directedness allows bidirectional mapping to our  social interaction representation as [context][policy][outcome], for combining social learning and intrinsic motivation. 
 
  It actively self-generates goals $\tau_g \in T$ by stochastically choosing the goals for which its empirical evaluation of learning progress is maximal. For each $\tau_g$, the robot explores  through goal-directed optimisation which policy $\pi_\theta$ can induce the given goal $\tau_g$ in context $c$. The exploration of the policy parameter space provides data to improve its estimation of the local forward model  $\mathbf{L}: (c, \theta) \mapsto \tau$ and inverse model $\mathbf{L^{-1}}:(c, \tau) \mapsto \theta $, that it can use later on to reach other goals.
 This autonomous exploration strategy is only interrupted when the teacher gives a demonstration $[c_{d},\zeta_{d},\tau_{d}]$, when it switches to the social learning strategy (fig. \ref{DataFlow}b). 
 
 With the social learning strategy, our SGIM-D learner imitates the demonstrated policy for a short while, and memorises the demonstrated outcome/goal as interesting, before resuming its autonomous exploration. It then generates a new goal, taking into account all its history, autonomous and social exploration phases alike. It chooses a goal with the highest interest level, which is defined as the competence progress.
 
The SGIM-D learner would thus try to explore goals  where it makes progress the fastest.  For each goal that it deems interesting, it would try different policies to approach it, using the policy repertoire of its past autonomous exploration or the policies suggested by the teacher's demonstrations.
 Once its competence for these easy goals is high, it no longer makes progress, and as its interest level for them drops, it progressively aims at more difficult goals and expands its search in the task space. The human teacher boosts its learning by indicating  policies to perform, so that its competence level increases, but also by indicating interesting goals/outcomes to emulate, to orient its search in the task space.
 
 In order to explain the design of the SGIM-D algorithm and before going into the details in section 3, we first consider a broader framework than that stated earlier in this section, to examine the different types of social interaction  in the literature before specifying the one used in this study.

\subsection{Analysis of Social Interaction Modes}

We would like to formalise the guidance of a human teacher to boost the learning of the relationship between the outcome $\mathfrak{T} \in \mathbb{T}$  and the policy $\pi \in \mathbb{P}$ in contexts $\mathfrak{C} \in \mathbb{C}$.

As in many approaches and for the sake of clarity, we assume in this section that the correspondence problem is solved, and do not differentiate the state, outcome and policy spaces between the robot and teacher.

Nevertheless, the two agents have acquired different knowledge, which changes throughout their interaction. We can describe this interaction as the way information flows between the human and the robot, intentionally or unintentionally:
\begin{itemize}
\item the human teacher's behaviour  or information flow from the human to the robot, $si_H$.
\item the robot learner's behaviour  or information flow from the robot to the teacher, $si_R$.
\end{itemize}

In order to define the social interaction that we wish to consider, let us characterise the different possibilities of information flow  as reviewed in \citep{Argall2009RAS,Billard2007RobotProgrammingby,Schaal2003PTRSLSBS,Lopes2009AbstractionLevelsfor} with respect to: what, how, when and who to imitate (see \citep{Dautenhahn2002, breazeal2002robots}. 
In this study, we only examine the possibilities of the information flow from the human to the robot $si_H$. Intentional communication from the robot to the human is a fundamental aspect of social learning \citep{Chernova2009JAIR,Thomaz2006}, and should be studied in a more extensive way in future work.

\subsubsection{What?}
Let us  examine the target of the information given by the teacher, or mathematically speaking, the space on which he operates. This can be either the policy, context or task spaces, or combinations of them.

\paragraph{Policy Space:}
Many social learning studies target the policy space $\mathbb{P}$. For instance, in programming by demonstration (PbD), $si_H$ shows the right policy to perform in order to reach a given goal. As an illustration, when teaching how to play tennis, your coach could show you how to hit a backhand by a demonstration, or by taking your hand and directing your movement. This approach relates to two levels of social learning: \textit{mimicry}, in which the learner copies the policies of others without an appreciation of their purpose, and \textit{imitation}, in which the learner reproduces the policies and the outcomes, as formalised in \citep{Lopes2009AbstractionLevelsfor,Call2002Threesourcesof,Whiten2000CS}.
   The policies demonstrated can be mimicked faithfully \citep{Cakmak2009IISRHIC}, be saved as corrections for the current situation \citep{Chernova2009JAIR}, form an initial dataset on which to build upon more complicated behaviour\citep{Argall2008,Argall2011RAS}, or indicate a locality to start an optimum search \citep{Peters2008NN}.  
     The information can be a trajectory or policy\citep{Peters2008NN},  high-level instructions\citep{Thomaz2006} or high-level advice\citep{Argall2008,Argall2011RAS}.  It can pertain to the entire policy, or only a part of it \citep{Argall2008,Argall2011RAS,Nicolescu2003,Thomaz2006}.
The literature often considers that targeting the policy space is the most directive and efficient method. However, it relies on the human teacher's expertise, which bears limitations such as ambiguity, imprecision, under-optimality or  the correspondence problem. Furthermore, the interaction is more effective at correcting visited situations, than exploring undemonstrated areas of  $ \mathbb{C}$ and $ \mathbb{T}$.

\paragraph{Context Space:}
The teacher can show interesting contexts $\mathfrak{C} \in \mathbb{C}$ in which the learner will have to work out.  To illustrate,  your tennis coach could train you specifically for situations where you are near the baseline while the ball falls near the net. Your coach would create this situation for you to handle, without saying which policy to perform. During infant-parent  joint play with toys, parents are able to play a role in the selection of the attended objects in the highly cluttered environment.
These processes of visual selection are realised by  implicit or explicit “social cues”  like pointing or gaze-following \citep{Slater2006,Tomasello2007DS}. Such social learning are classified as \textit{stimulus enhancement} or \textit{observational conditioning}\citep{Whiten2000CS}. 
 The teacher can select objects to be attended to \citep{Cakmak2009IISRHIC}, structure the environment by defining landmark states \citep{Thomaz2006}, indicate desirability of contexts through reinforcement signals   \citep{Thomaz2008CS}, or give advice  \citep{Argall2008, Argall2011RAS}. 
 
 Whereas acting on the context space does not  speed up the learning progress, it enables the learner to explore new situations. 

\paragraph{Task Space:}
The third kind of information is about possible outcomes $\mathfrak{T} \in \mathbb{T}$, and is related to goal-directed exploration, where the learner focuses on discovering different outcomes instead of different means of completing the same goal. This pertains to the \textit{emulation} level of social learning, where the observer witnesses someone produce a result on an object, but then employs his own policy repertoire to reproduce the result, as formalised in \citep{Lopes2009AbstractionLevelsfor,Call2002Threesourcesof,Whiten2000CS,Nehaniv2007}. 
Your tennis coach could ask you to hit with the ball the right corner of the court, wherever you received the ball, whichever shot you use.
Goal-directed approaches  allow the teacher to reset goals \citep{Argall2008}, to request the execution of goals\citep{Thomaz2006} or to label goal states \citep{Thomaz2006, Thomaz2008CS}.  The learner can infer from the demonstrations the goal by positional and force profiles to iron and open doors \citep{Kormushev2011AR}, or by using inverse reinforcement learning\citep{Lopes2011IICDL}.
This approach is essential to learn multiple tasks/goals, and all the more interesting as it is inspired by psychological behaviours \citep{Whiten2000CS,Tomasello2007DS,Csibra2003PTRSLSBS}. The drawback is that the learning needs a policy repertoire large enough to be used to reach various goals, before it improves.

As we want the learner to accomplish not only a single goal but to be efficient on a large variety of goals, we choose to bootstrap its learning with information targeting the task space. Furthermore, we also want the learning process to benefit from the social interaction early. So that the learner builds its policy repertoire quickly, we choose to target the policy space $\Pi$ too.

\subsubsection{How?}
Whichever the target, the information can be communicated from the teacher to the learner in several ways:

\paragraph{Demonstration at a low level:}
 The teacher performs the policy or shows the context or goal\citep{Cakmak2009IISRHIC,Chernova2009JAIR,Peters2008NN} : the information flow $si_H \in \mathbb{C} \cup \mathbb{P} \cup \mathbb{T}$.
This approach is the most natural for non-expert teachers, and requires little training for the teacher. However, demonstrations are generally assumed of high quality, whereas in reality, they can be ambiguous, unsuccessful or suboptimal in certain areas. 

\paragraph{Demonstration at a high level:}
 The teacher shows the context/policy/goal at a symbolic level. A language protocol often enables instructions of policies  \citep{Nicolescu2003,Thomaz2006,Thomaz2008CS,Argall2008,Argall2011RAS}, or suggestions of goals \citep{Thomaz2006,Thomaz2008CS}. In this case, $si_H \in \tilde{\mathbb{C}}$ or $\tilde{\mathbb{P}}$ or $\tilde{\mathbb{T}}$, which bear a direct transformation to $\mathbb{C,P}$ and $\mathbb{T}$.
A high-level approach seems more natural by the use of a language, but it is dependant on the predefined communication channel and often lacks flexibility for new situations or changing environments. It also entails a training before  the teacher can efficiently communicate with the robot.

\paragraph{Advice:}
 The teacher shows the desired context/policy/goal indirectly. He does not show the right desired state but indicates how to approach that state\citep{Argall2008,Argall2011RAS}. $si_H$ is a function of the context/policy/goal experienced by the robot and the desired value.
 Advice is an efficient way of providing instructions at a high-level even for continuous environments, while avoiding the limitations of the demonstrator's performance, as well as the re-creation of difficult or dangerous states. Nevertheless advice is an indirect way of giving instructions, which may be imprecise and limited by the language definition, which again lacks flexibility and requires the teacher adapting to it.

\paragraph{Reward:}
 Reward-like signals ($si_H \in \mathbf{R}$) or  "good or bad" indications ($si_H \in \{-1;1\}$) are common in reinforcement-based approaches, which  benefit considerably from the formalism of reinforcement learning \citep{Nicolescu2003,Thomaz2006,Thomaz2008CS}. 
They easily couple social learning with techniques of learning from experience. However, defining the reward function is known to be non-trivial. Especially, human teachers tend to give anticipatory and asymmetrically positive rewards \citep{Thomaz2006}. Taking into account the non-Markovian behaviour of human beings would induce high complexity in the reinforcement learning framework. Furthermore, reinforcement learning research has so far focused on reaching a single goal $\mathfrak{T} \in \mathbb{T}$, and not a set of goals.

\paragraph{Labelling:}
 A few works have labelled previously reached goals to help structure the environment and facilitate communication between the teacher and the learner  \citep{Thomaz2006,Thomaz2008CS}. In this case, $si_H$ takes discrete values that symbolise the different classes.
 
As we wish in the model and experiments presented below, to address the learning for large, complex and continuous environments, so that the robot learns a wide variety of goals/tasks, we opt for low-level demonstrations. So as 
to minimise the correspondence issues, we teleoperate our robot using kinaesthetic demonstrations, while recording from its own sensors.

\subsubsection{When?}
The timing of the interaction varies with respect to its timing within an episode [context][policy][outcome], and with respect to its general activity during the whole learning process.
\paragraph{Timing within an activity episode:}
If we consider that each activity episode involves a reading of the context state of the environment, before performing a policy, and finishes by observing the outcome in the environment, we can classify the various types of timing of the interaction into two types:
\begin{itemize}
\item{Feedback:} A past-directed message informs the learner about its past behaviour.
The chronology would be [$\mathfrak{C}$][$\pi$][$si_H$][$\mathfrak{T}$] or  [$\mathfrak{C}$][$\pi$][$\mathfrak{C}$][$si_H$]. These messages can be good/bad assessments on its past behaviour \citep{Thomaz2006,Thomaz2008CS,Nicolescu2003,Lopes2011IICDL},  a scalar reward given by the human teacher \citep{Thomaz2006}, a  correction demonstration \citep{Chernova2009JAIR}, an advice to modulate the wrong behaviour \citep{Argall2008,Argall2011RAS}, or a label of previously reached goals \citep{Thomaz2006,Thomaz2008CS}.
According to his partial knowledge of the internal state of learning of the robot, the human adapts his teaching. However, the robot trial policy can be time consuming when it is  very far from any good solution.

\item{Feedforward:} A future-directed message informs the learner before deciding its future behaviour. 
The chronology would be [$\mathfrak{C}$][$si_H$][$\pi$][$\mathfrak{T}$]. These messages are commonly instructive demonstrations of good example behaviours\citep{Cakmak2009IISRHIC,Chernova2009JAIR}.   Not only have behavioural studies shown that human teachers tend to give future-directed messages \citep{Thomaz2006}, feedforward messages also seem more instructive with respect to the immediate future behaviour of the robot. However they do not take into account any information flow from the robot to the teacher.
\end{itemize}

\paragraph{General timing during the whole learning process:}
The rhythm of social interaction varies considerably among studies of social learning:
\begin{itemize}
\item{At a fixed frequency:} In classical imitation learning, the learner uses a demonstration to improve its learning at every policy it performs\citep{Argall2008,Argall2011RAS,Cakmak2009IISRHIC}. 
This solution is ill-adapted to the teacher's availability or the needs of the learner, who requires more support in difficult situations. Though, this continuous interaction allows steady bootstrapping of the learning and adaptation to changing environments.

\item{Beginning of learning:} A limited number of examples is given to initialise the learning, as a basic behaviours repertoire \citep{Argall2008,Argall2011RAS}, or  a sample behaviour to be optimised\citep{Peters2008NN,Kormushev2010IROS}. The learner is endowed with some basic competence before self-exploration. Nevertheless, if the interactions are restricted to the beginning, the learner could face difficulties adapting to changes in the environment.

\item{At the teacher's initiative:} The teacher alone decides when he interacts with the robot\citep{Thomaz2006}. In most examples, the teacher gives corrections when seeing errors \citep{Koenig2010NN,Cakmak2010AMDIT}, to restrict human interventions to when it is needed. Nevertheless, it still is time consuming as he needs to monitor the robot's errors to give adequate information to the learner.

\item{At the learner's initiative:} The learner can request for the teacher's help in an ambiguous \citep{Chernova2009JAIR,Cakmak2010AMDIT}  or unknown\citep{Thomaz2006} situation, or only reproduces the observations when it observes an outcome that matches its goal  during goal-based imitation or mimicking\citep{Cakmak2009IISRHIC}. This approach is the most beneficial to the learner, for the information arrives as it needs them, and the teacher needs not monitor the process. 
\end{itemize}
 
 These 4 types can be classified into 2 larger groups: 
 \begin{itemize}
 \item batch learning, where the data provided to the learner is decided before the learning phase, and is given independently of the learning progress, generally in the beginning of the learning phase.
 \item interactive learning, where  the user interacts with the incrementally learning robot, either at the teacher's or the learner's initiative.
 \end{itemize}
 
 \subsubsection{Chosen Approach}
In the model and experiments presented below, we choose to use a feedforward signal, as it is more natural for human teachers. For simplicity reasons, we set the interaction at regular frequency, allowing easier assessment of our SGIM-D algorithm and comparison with other learning algorithms. 
 In this proof-of-concept study, we deliberately ignored the who question, which examines  cases of multiple teachers. This very stimulating question yet requires a separate examination to avoid too much complexity in a single study.
 Among this listing of social learning, our choice of information flow is:
 \begin{itemize}
 \item{What:} We opted for an information flow targeting both policy and task spaces,  to enable the biggest progress for  the learner. 
 \item{How:}  We will be giving low-level demonstrations of possible policies and goals through kinaesthetic demonstration, for this seems the most efficient to teach at the low level to enable to work in real-world continuous and changing environments. This choice avoids any symbolic thus discrete representations of policies or the environment, or a preset language to communicate at the high-level.
 \item{When:} Although interactive learning  at either the learner's or the teacher's initiative seems interesting theoretically, it introduces combinatorially many variants. To evaluate the basis of our architecture, we choose to trigger the interaction at a regular frequency.
 \end{itemize}

Having decided the specifications for our algorithm with respect to the social interaction mode and the autonomous exploration algorithm, let us detail its structure in the next section.

\section{SGIM-D Algorithm}

\begin{figure*}
\centering
\includegraphics[width=\textwidth]{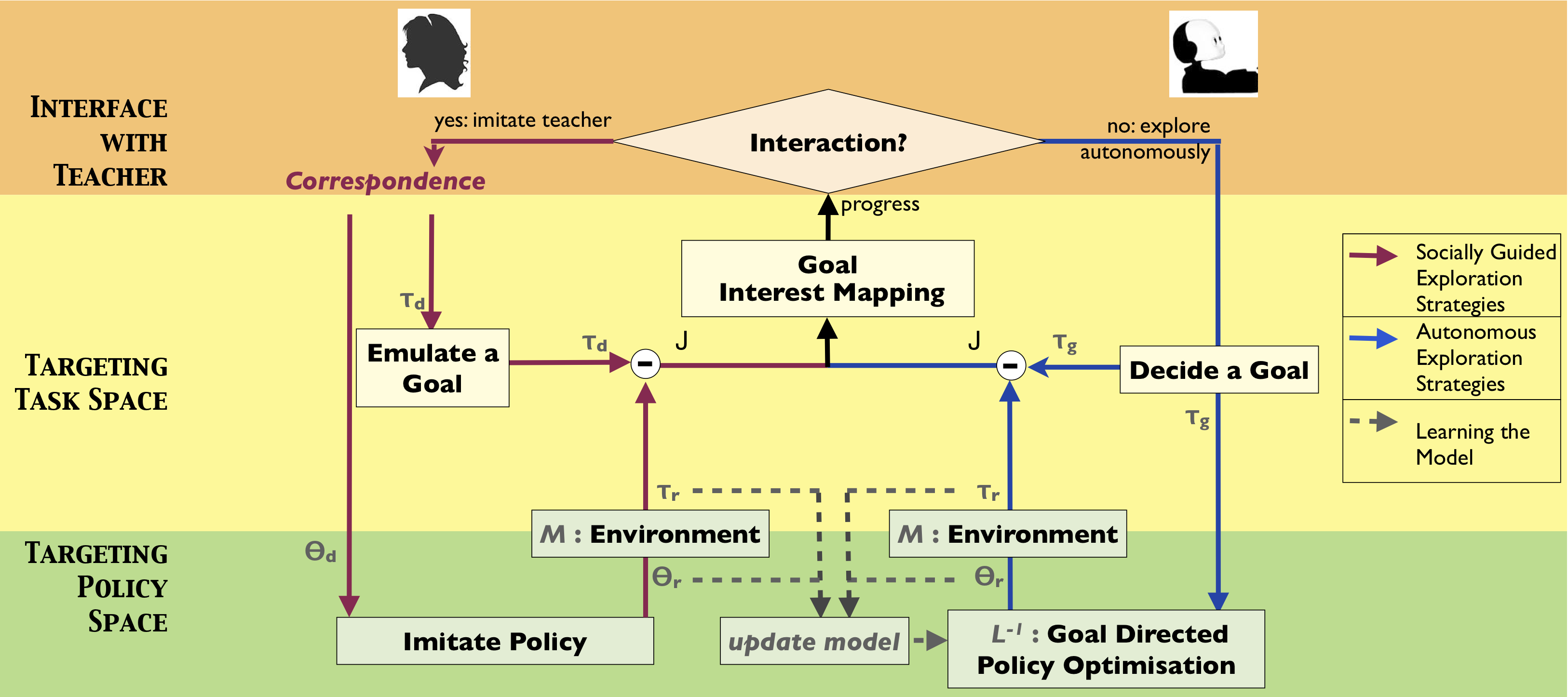}
\caption{Time flow chart of SGIM-D  into 3 layers that pertain to the human-machine interface, the task space exploration and the policy space exploration respectively. The architecture combines sub-modules for intrinsically motivated learning and socially guided learning on both the policy and task levels.}
\label{StructureSGIM}

\vspace{0.5cm}

\input{PseudoCodeSGIMD.tex}

\end{figure*}

Socially Guided Intrinsic Motivation by Demonstration (\textbf{SGIM-D}) is an algorithm that merges  programming by demonstration as social interaction strategy with the SAGG-RIAC algorithm as intrinsic motivation strategy, for the learning of local inverse and forward models in complex, high-dimensional and continuous spaces.
Its architecture is separated into three layers where sub-modules of each strategy interact (fig. \ref{StructureSGIM} and  algo. \ref{alg:PseudoCodeSGIMD}) :
\begin{itemize}
\item An interface with the teacher, which manages the "physical" interaction with the teacher. It detects that the teacher performs a demonstration and translates it into parameters for the robot. The implementation of this interface is specific to each robot and experimental setting, and will be detailed specifically for the experimental setup in section \ref{CorrespondenceMapping}.
\item A higher level of active learning, the \textit{Task Space Exploration level}  which drives the exploration of the task space. It sets goals $\tau_g$ depending on their interest levels that is based on the competence of previous goals, retrieves from the teacher information about goals, and maps $T$ in terms of interest level. It learns at a longer time scale. Its structure is detailed in subsection \ref{Goal Space Exploration Level}.
\item A lower level of active learning, the \textit{Policy Space Exploration level} that explores the policy parameter space $\Pi$ to improve its estimation of $J$ and estimate the inverse mapping $\mathbf{L^{-1}}$. While interacting with the teacher, it would imitate his policies $\zeta_{d}$, whereas during self-exploration, it would attempt to reach the goals $\tau_g$ set by the Task Space Exploration level. It  learns at a shorter time scale. Its structure is shortly described in subsection \ref{Policy Space Exploration Level} and detailed for our implementation in section \ref{Fishing Experiment}.
\end{itemize}

\subsection{Lower Level : Policy Space Exploration}
\label{Policy Space Exploration Level}

The \textit{Policy Space Exploration} searches the policy parameters space $\Pi$ to reach the goal $\tau_g$ set by the higher level or imitates the demonstrated policy $\zeta_{d}$, and returns to the Task Space Exploration level the measure of competence at reaching  $\tau_g$.

The implementation details will depend on the experimental setup, but mainly, the Policy Space Exploration level  contains 2 functions:
\begin{itemize}

\item The \textit {Imitate Policy} function takes as input a policy parameter $\theta_{d}$ demonstrated by the teacher and tries to repeat it. This function can be changed to match other social interaction behaviours. An implementation is described for our experimental setting in subsection \ref{Imitate Policy}.

\item The \textit{Goal-Directed Policy Optimisation} function  searches for policy parameters $\theta$ that guide the system toward the goal $\tau_g$ in the given context $c$ by 1) building local inverse $\mathbf{L^{-1}}$  model during exploration that can be re-used for later goals and 2) selecting new policies depending on interestingness measures of policies with respect to the current goal to get a better estimate of $J$. Mainly, it can be implemented by classical autonomous learning methods mentioned earlier which learn for a single goal only such as classical reinforcement learning methods. An example is presented for our experimental setting in subsection \ref{Decide a Policy}.
 This function optimises $\theta \mapsto J(\tau_g, M(c, \theta))$.
 
\end{itemize}

\subsection{Higher Level : Active Goal Babbling for Task Space Exploration}
\label{Goal Space Exploration Level}
The \textit{Task Space Exploration} relies on feedback from  the Policy Space Exploration level to decide which goal $\tau_g \in T$ is interesting  to focus on. It explores $T$  using teacher's demonstrations of goals (\textit{Emulate a Goal}) and self-determines a goal (\textit{Decide a Goal}) using competence measures, more precisely competence improvement it maps on $C \times T$ (\textit{Goal Interest Mapping}).

\subsubsection{Goal Interest Mapping Function}

To determine which goals it should attempt to better generalise for the whole task space, we assign a competence $\gamma_{c,\tau_g}$ to each task $\tau_g$ explored in context $c$,   as a measure of how close the learner can reach $\tau_g$:

\begin{eqnarray}
\gamma_{c,\tau_g}=  min_{(c,\theta,\tau_g) \in Memo} J(\tau,M(c,\theta))
\label{competence}
\end{eqnarray}

 where $Memo$ is the list of all the past episodes $(c,\theta,\tau)$.
 
 Along with the estimated inverse mapping function $\mathbf{L^{-1}}$, SGIM-D estimates at the same time the interest mapping function over $C \times T$(algo. \ref{alg:PseudoCodeSGIMD}, l. \ref{line:goalInterestMapping})). In our approach, while $\mathbf{L^{-1}}$ is estimated as a complex function, we model the interest mapping as a piecewise constant function.

Let us consider a partition $\biguplus_i {R}_i = C \times T$. 
Each ${R}_i$ contains attempted goals given a context $\{(c_{t_1}\tau_{t_1}), (c_{t_2},\tau_{t_2}), ...$ $(c_{t_k}, \tau_{t_k})\}_{{R}_i}$ of competences $\{\gamma_{{t_1}}, \gamma_{{t_2}}, ..., \gamma_{{t_k}}\}_{{R}_i}$, indexed by their relative time order of experimentation $t_1< t_2< ...< t_k  $ inside subspace ${R}_i$. 

An estimation of interest is computed for each region $R_i$ as \textit{ the local competence progress, over a sliding time window of the $\mathbf{\zeta}$ more recent goals attempted inside ${R}_i$}:

\begin{center}
\vspace{-0.4cm}
\begin{eqnarray}
interest_i  =  \frac{ \left| \left(\displaystyle \sum_{j=| {R}_i|-\zeta}^{|{R}_i|-\frac{\zeta}{2}} \gamma_{j} \right) - \left(\displaystyle \sum_{j=|{R}_i|-\frac{\zeta}{2}}^{|{R}_i|} \gamma_{j} \right) \right|}{\zeta} 
\label{interest}
\end{eqnarray}
\end{center}

By using a derivative, the interest considers the \emph{variation of competences}, and by using an absolute value, it considers cases of \emph{increasing and decreasing competences}. In SGIM-D, we will use the term \textit{competence progress} with its general meaning to denote this increase and decrease of competences.
An increasing competence signifies that the expected competence gain in $R_i$ is important. Therefore, selecting new goals in regions of high competence progress could bring both a high information gain for the learned model, and also drive the reaching of previously unachieved goals. 
Depending on the starting position and potential evolution of the environment, a decrease of competences inside already well-reached regions can arise. In this case, the system should be able to focus again in these regions to attempt to re-establish a high level of competence inside. This explains the usefulness of considering the absolute value of the competence progress as shown in equation \ref{interest}.

Based on this definition of interest, the module builds an interest level mapping, at each new goal $\tau_g$ explored by autonomous exploration or at each goal $\tau_{d}$ observed in social learning. It partitions $C \times T$ into subspaces, so as to maximally discriminate areas according to their levels of interest, as described in \citep{Baranes2013RAS}.
We use a recursive split of the space, each split occurring once a maximal number of goals have been attempted inside. Each split maximises the difference of the \textit{interest} measure in the two resulting subspaces, and easily separates areas of different interest, and thus, of different reaching difficulty (cf. algorithm  \ref{alg:UpdateRegions}).

\begin{figure*}
\input{PseudoCodeUpdateRegions.tex}
\end{figure*}

\subsubsection{Decide a Goal Function}

The  \textit{Decide a Goal} function uses the interest level mapping to select the next goal to perform   (algo. \ref{alg:PseudoCodeSGIMD}, l. \ref{line:decideGoal})).
Goals are chosen stochastically according to either of the following modes:
\begin{itemize}

\item \textbf{Mode(1)}:  A chosen random goal inside a region which is selected with a probability proportional to its interest value. The probability of selecting the region ${R}_n$ that contains the current context $c$ is: 
\vspace{-0.4cm}

\begin{eqnarray}
P_n = \frac{ interest_n - \textbf{min}(interest_i)}{\sum_{i=1}^{|R_n|}interest_i - \textbf{min}(interest_i)}
 \label{goalSelection}
\end{eqnarray}
\item \textbf{Mode(2)}: A selected random goal inside the whole space $T$. \\
\item \textbf{Mode(3)}:  A first selected region according to the interest value (like in $mode(1)$) and then a generated new goal close to the already experimented one which received the lowest competence estimation $min_{R_n} (\gamma_{t_i})$.

\end{itemize}

\subsubsection{Emulate a Goal Function}

At each demonstration, the learner observes not only the policy performed, but also its outcome $\tau_{d}$. It henceforward considers this outcome as a potential goal, and assigns an interest level according to its own policy repertoire and model it has built   (algo. \ref{alg:PseudoCodeSGIMD}, l. \ref{line:emulate})).

The above description is detailed for SGIM-D's choice of imitating teachers' low-level demonstrations of outcomes and policies. Such a structure would remain suitable for other choices of social interaction modes, and we only have to change the content of the \textit{Emulate a Goal} function, and change the \textit{Imitate a Policy} function to match the chosen behaviour.

In the following section, we illustrate the principle of SGIM-D through a proof-of-concept experiment, where our robot will learn how to fish.

\section{A Case Study: the Fishing Rod Environment}
\label{Fishing Experiment}

In this section we describe our experimental setup with the environment's description, and then detail how SGIM-D functions adapt for this specific setup.

In this illustration experiment, we consider a simulated  6 degrees-of-freedom robotic arm holding a fishing rod (fig. \ref{FishingRod}). The aim is that it learns how to reach any point  on the surface of the water with the float at the tip of the fishing line.   This is an inverse model in a continuous and unbounded environment of a complex system that can hardly be described by physical equations.

In our experiment, the context space $C$ describes the initial actuator/joint positions and state of the fishing rod. $T=[-1,1]^2$ is a 2-D space that describes the position of the float when it reaches the water. For each position $\tau \in T$, it has to learn a new goal : with which movement $\pi_\theta$ he can place the float closest to $\tau$. The robot base is positioned at (0,0) and it always starts with the same configuration $c_{org}$.

\subsection{Motor Primitives and Correspondence Mapping}
\label{CorrespondenceMapping}

Variable $\theta$ describes the parameters of the motor primitives of the joints, defining for each joint 4 scalar parameters that represent the joint positions at $t=0$, $t=\frac{\delta}{3}$,$t=\frac{2 \delta}{3}$ and $t=\delta$.  These 4 parameters $u_1,u_2,u_3, u_4$ generate a trajectory for the joint by Gaussian distance weighting:

\begin{eqnarray}
 u(\mathbf{t}) = \sum_{i = 0}^{4}{ \frac{ w_i(\mathbf{t}) u_i } { \sum_{j = 0}^{4}{ w_j(\mathbf{t}) } } }
  \mbox{ with }  
  w_i(\mathbf{t}) =  e^{\sigma* |\mathbf{t} -\mathbf{\frac{i \delta}{3}}|^2} , \sigma>0
  \label{gdw}
\end{eqnarray}

 Each of the 6 joints' trajectories is determined by 4 parameters. Another parameter sets $\delta$. Therefore a policy is represented by $ 6 \times 4 + 1= 25$  parameters: $\theta= (\theta^1,\theta^2, ...\theta^{25} )$.  $\Pi = [0,1]^{25}$. This choice of taking only 4 samples of the movement trajectory is arbitrarily, and other parametrisations have been also used in other studies  \citep{Nguyen2012PPCRDC,Nguyen2012IICHR}.
 
 Because our experiment uses for each trial the same context $c_{org}$, our system memorises after executing  every policy parameter $\theta$, simply the context-free association $ \theta \mapsto \tau$.
 
Upon observation of a demonstration $(\zeta_{Hd}, \tau_{Hd})$, the \textit{Correspondence} function first computes the parameters $\theta_{d}$ that enable him to reproduce the teacher's policy $\zeta_{d}$ closest (algo. \ref{alg:PseudoCodeSGIMD}, l. \ref{line:correspondance})).

\begin{figure}
\centering
\includegraphics[width=0.5\textwidth]{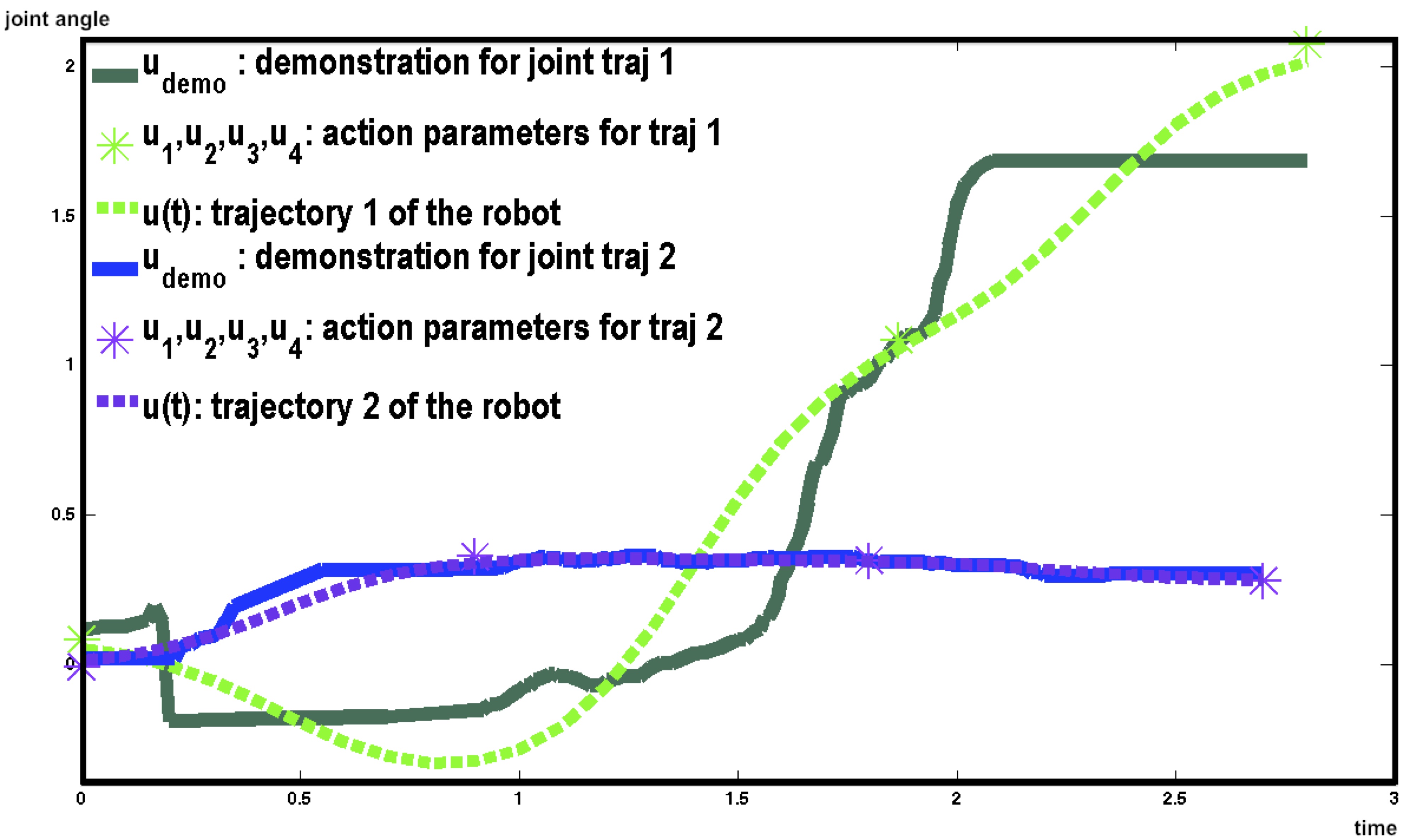}
\caption{ Mapping of the demonstrations given by the human teacher by the robot. Horizontal axis: time, vertical axis: joint angle (best seen in colors). Are plotted for 2 different joint trajectories of a human demonstrator, the demonstrated trajectory, and the corresponding movement parameters and trajectory mapped by the robot. For a demonstrated trajectory $u_{Hd}$, parameters $u_1, u_2, u_3, u_4$ minimise eq. \ref{demonstrationsMappingEq}. Then the parameters $u_1, u_2, u_3, u_4$ generate the trajectory executed by the robot $u(t)$ according to eq. \ref{gdw}. 
 For joint trajectory 1, the mapping has a high error value, while for joint trajectory 2, the mapping has a low error value.\\
}
\label{demonstrationsMapping}
\end{figure}

From $\zeta_{Hd}$, it can extract for each joint a trajectory $u_{Hd}(t)$ and the duration of the trajectory $\delta$.
To map a given joint trajectory $u_{Hd}(t)$ into our robot's parameterised dynamic motor primitive, we need to determine the 4 parameters $u_1,u_2, u_3 , u_4$  so as to minimise the error (fig. \ref{demonstrationsMapping}):

\begin{eqnarray}
d = \left| \left| u_{Hd}(\mathbf{t}) - \sum_{i = 0}^{4}{ \frac{ w_i(\mathbf{t}) u_i } { \sum_{j = 0}^{4}{ w_j(\mathbf{t}) } } } \right| \right|_2 
\label{demonstrationsMappingEq}
\end{eqnarray}

$\theta_{d}$ is thus the set of parameters  $u_1,u_2, u_3 , u_4$  for each of the 6 joints, and $\delta$, which minimise $d$ by the trust-region-reflective algorithm described in \citep{Coleman1994MP,Coleman1996SJO}.

\subsection{Imitate a Policy}
\label{Imitate Policy}
  The \textit{Imitate a Policy} function  (cf. algorithm  \ref{alg:PseudoCodeImitation} and fig. \ref{StructureSGIM}) explores the locality of $\theta_{d}$ with policy parameters  $\theta_{imitate} = \theta_{d} + \theta_{rand} $  (  algo. \ref{alg:PseudoCodeImitation}, l. \ref{line:randImitate}) with $\theta_{rand}$ a random movement parameter variation, so that $|\theta_{rand}|< \epsilon$. After a short fixed number of times,  SGIM-D computes its competence at reaching the goal indicated by the teacher $\tau_{d}$  (cf. eq. \ref{competence}), then, shifts back to the autonomous exploration mode. The measure of competence returned  is defined hereafter.

\begin{figure}
\input{PseudoCodeImitation.tex}
\end{figure}

\begin{figure}
\input{PseudoCodeExecute.tex}
\end{figure}

\subsection{Performance Measure}
\label{Sim}
We define J as the euclidian distance  $D(\tau_g,\tau)$, and normalised by the distance between the original position $\tau_{org}$ and the goal: $ D(\tau_{org}, \tau_g)$. This allows, for instance, to give the same competence level when considering a goal at 1km from the origin position that the robot approaches at 0.1km, and a goal at 100m that the robot approaches at 10m:

\vspace{-0.5cm}
\begin{eqnarray}
J(\tau_g, \tau) = \left\{
\begin{array}{ll}
 -1 &  \mbox{if}  \frac{D(\tau,\tau_g)}{D(\tau_g, \tau_{org})} > 1\\
 \\
   -  \frac{D(\tau,\tau_g)}{D(\tau_g, \tau_{org})}    &  \mbox{otherwise}  
 \end{array}
 \right.
\end{eqnarray}
 \label{similarity}
\vspace{-0.4cm}

Here, our direct model $M :\theta \mapsto \tau$ only considers the 25 parameters $\theta= (\theta^1,\theta^2, ...\theta^{25} )$ as inputs of the system, and a position in $\tau = (\tau^1,\tau^2)$ as output. We wish to build the estimate inverse model $L^{-1} :  \tau \mapsto \theta $ by using the following optimisation mechanism for goal-directed exploration and learning, which can be divided into two different strategies.
 
\subsection{Goal-Directed Policy Optimisation}
\label{Decide a Policy}
The \textit{Goal-Directed Policy Optimisation} function (cf. algorithm  \ref{alg:PseudoCodeExploration} and fig. \ref{StructureSGIM}) learns to reach a goal generated by the Task Space Exploration level. This function can be implemented by any single task learning algorithm.
For the sake of proving that the efficiency of our SGIM-D algorithm relies on its  general structure, and not so much on its per-goal learning algorithm,  we choose an exploration method that builds memory-based local direct and inverse models, balancing between global exploration and local optimisation to avoid local minima.
To decide which mode is triggered given a goal $\tau_g$, we examine the memory of the system, and consider that the closest one has been able to reach $\tau_g$, the more the system should focus on local optimisation. On the contrary, if during the system's history, it has never reached a point close enough to the goal $\tau_g$, it should prefer global exploration.

The system continuously estimates the distance between the goal $\tau_g$ and the closest already reached position $\tau_c$: $J(\tau_c, \tau_g)$. The system has a probability proportional to  $J(\tau_c, \tau_g)$ of being in the Global Exploration regime, and the complementary probability of being in the Local Optimisation regime.

\begin{figure}
\input{PseudoCodeExploration.tex}
\end{figure}

\subsubsection{Global Exploration Regime}
In this regime the system just picks random policy parameters  $\theta \in T$ to explore the policy space (algo.  \ref{alg:PseudoCodeExploration},  l. \ref{line:randomMovement}).

\subsubsection{Local Optimisation Regime}
The local optimisation regime  (algo.  \ref{alg:PseudoCodeExploration},  l. \ref{line:localOptimisation}) uses the memory data to infer locally inverse models $ L^{-1}: (\tau^1,\tau^2) \rightarrow  (\theta^1,\theta^2, ...\theta^{25} ) $.  Given the high redundancy of the problem, we choose a local approach and extract the potentially more reliable data using the following method (algo. \ref{alg:LocalData}). First (algo. \ref{alg:LocalData}, l. \ref{line:localDataH}), we compute the set $H$ of the $h_{max}$ nearest neighbours of $\tau_g$ and their corresponding movement parameters using an ANN method \citep{Muja09}, which is based on a tree split using the k-means process :
\begin{eqnarray}
H &=&  \left\{  (\tau,\theta)_1,  (\tau,\theta)_2,   ... ,  (\tau,\theta)_{h_{max}}   \right\}   \subset (T\times \Pi)^{h_{max}} 
\end{eqnarray}
Then, for each element $(\tau,\theta)_h \in H$, we compute its reliability. Let us consider the set $K_h$ which contains the nearest neighbours of $\theta_h$ within $distN$ of $\theta_h$ in the memory set \textbf{M} with respect to norm $||.||_2$ (algo. \ref{alg:LocalData}, l. \ref{line:localDataK}) :
\begin{eqnarray}
K_h &=& \left\{  (\tau,\theta)_1,  (\tau,\theta)_2,   ... ,  (\tau,\theta)_{k_{max}}   \right\}
\end{eqnarray}
As the reliability of the local model depends both on the knowledge of the locality and the reproducibility of the movement due to non-linear noise that produces small variations in $\tau$ of magnitude depending on $\theta$ (fig. \ref{deterministic}), we define for each element $(\tau,\theta)_h \in H$, its reliability as $dist(\tau_h,\tau_g) + \alpha\times var_h$, where $var_l$ is the variance of the set $K_h$, and $\alpha $ is a constant set to 0.5 in our experiment. We choose the smallest value, as the most reliable set $(y,\theta)_{best}$ (algo. \ref{alg:LocalData}, l.  \ref{line:localDataR}).

\begin{figure*}
\input{PseudoCodeLocalData.tex}
\end{figure*}

In the locality of the set $(\tau,\theta)_{best}$,  we interpolate using the $k_{max}$ elements of $K_{best}$ to compute the policy corresponding to $\tau_g$ :
$\theta_g  = \sum_{k=1}^{k_{max}} {\beta_k \theta_k} $ where  $\beta_k \propto Gaussian(dist(\tau_k,\tau_g)) $ is a normalised Gaussian of the euclidian distance between $\tau_k$ and the goal $\tau_g$.

We execute policy of parameter $\theta_g$ and continue with the Nelder-Mead simplex algorithm \citep{Lagarias1998SJO}, to minimise the distance of the outcome $\tau_2$ to the goal $\tau_g$. This algorithm uses a simplex of 26 points for 25-dimensional vectors $\theta$. It first makes a simplex around the initial guess $\theta_g$ with the $\theta_k, k= 1,...k_{max}$. It then updates the simplex with points around the locality until the distance to minimise falls below a threshold.

\subsection{Stochastic Environment}

 \begin{figure}
\centering
\includegraphics[width=0.5\textwidth]{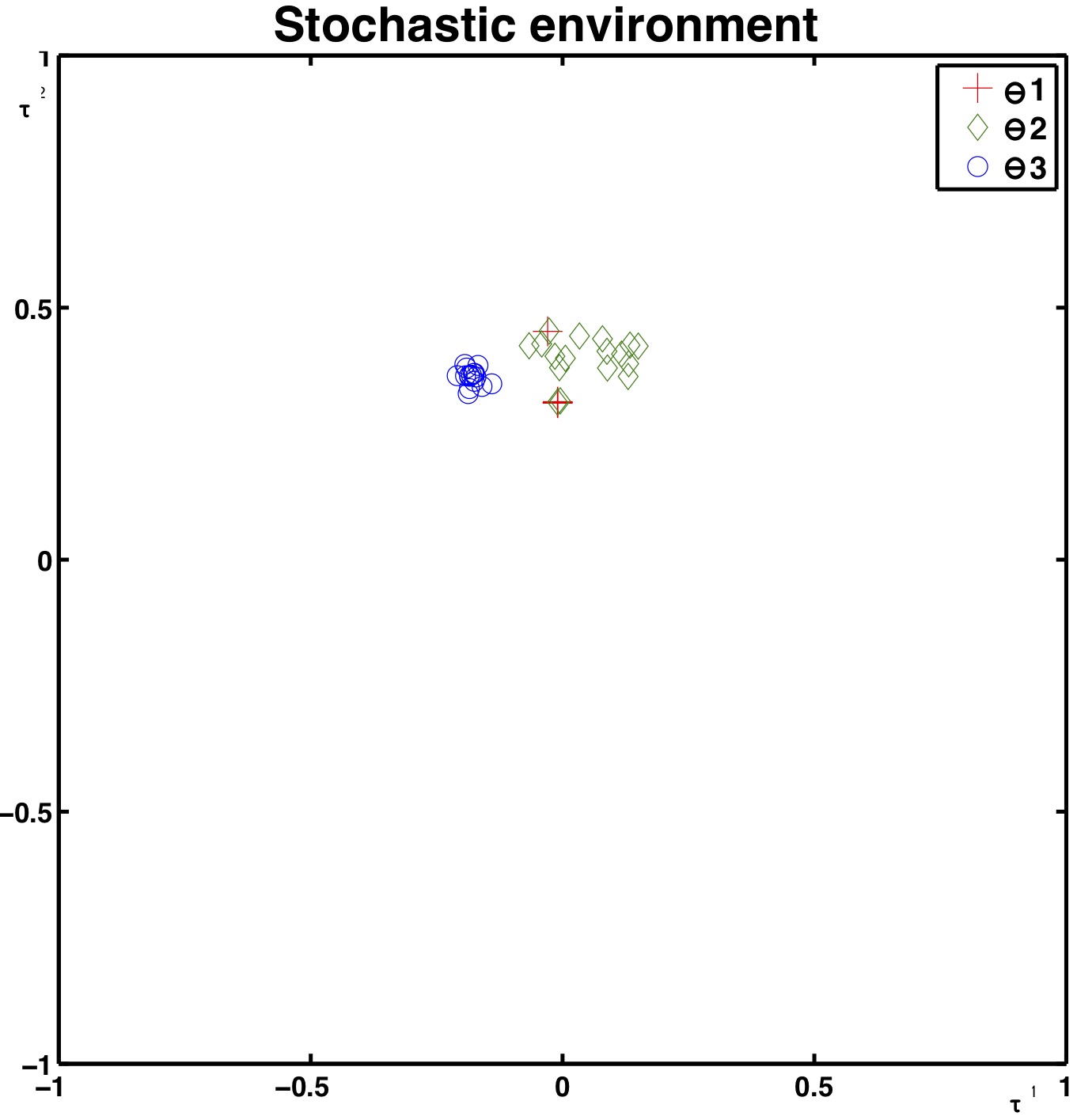}
\caption{ Outcomes for 3 different policy parameters  over 20 repetitions of the same movement, represented in the 2-D space $T$. Standard deviations are for each policy parameters, respectively (0.005, 0.033) for a1, (0.0716, 0.041) for $\theta2$, and (0.016, 0	.016) for $\theta3$ (best seen in colors).}
\label{deterministic}
\end{figure}

All the experimental setup has been designed for a 6 DOF robot arm in the real world. Nevertheless, to be able to collect statistics through numerous experiments, we built a model of our 6 DOF arm on V-REP physical simulator, which uses a ODE physics engine that updates every 50 ms.

 Due to stochasticity of the simulated experimental setup, repetitions of the same movement do not lead to the same exact outcome. 
Moreover, the stochasticity does not follow a uniform distribution rule and can not be modelled by a simple Gaussian. The standard deviation varies along the different dimensions and depends on the dynamic properties of the movement performed (fig \ref{deterministic}).
  The mean variance of the control system of the robot is estimated to 0.073 for  measures of 10 attempts  of  20 random policy parameters, while the reachable area spans between -1 and 1 for each dimension of $T$.

This  fishing experiment focuses on the learning of inverse models in a continuous space, and deals with high-dimensional and highly redundant models. Our setup is all the more interesting as a real-world fishing rod's and flexible line's dynamics would be difficult to model.  The model of a fishing rod in the simulator might be mathematically computed. However, To represent the complexity of the fishing line manipulated by the robot arm, we modelled it as a set of 30 segments and 30 revolute joints, which leads to complex movements hard to predict. Even though the direct mapping has been modelled by the simulator,  the inverse model, which is even more complicated due to redundancy and stochasticity, is yet to learn. Besides, our fishing environment's stochasticity distribution is hard to model.  Thus learning directly the outcome of one's policies is all the more advantageous.

The next section describes how  we evaluate the SGIM-D algorithm using the fishing experimental setup.

\section{Experimental Protocol}
In this section, we detail the experiments we carry with our fishing robot setup to evaluate SGIM-D and  how we provide our learner with demonstrations.
\subsection{Comparison of Learning Algorithms}

\begin{figure*}
\centering
\includegraphics[width= \textwidth]{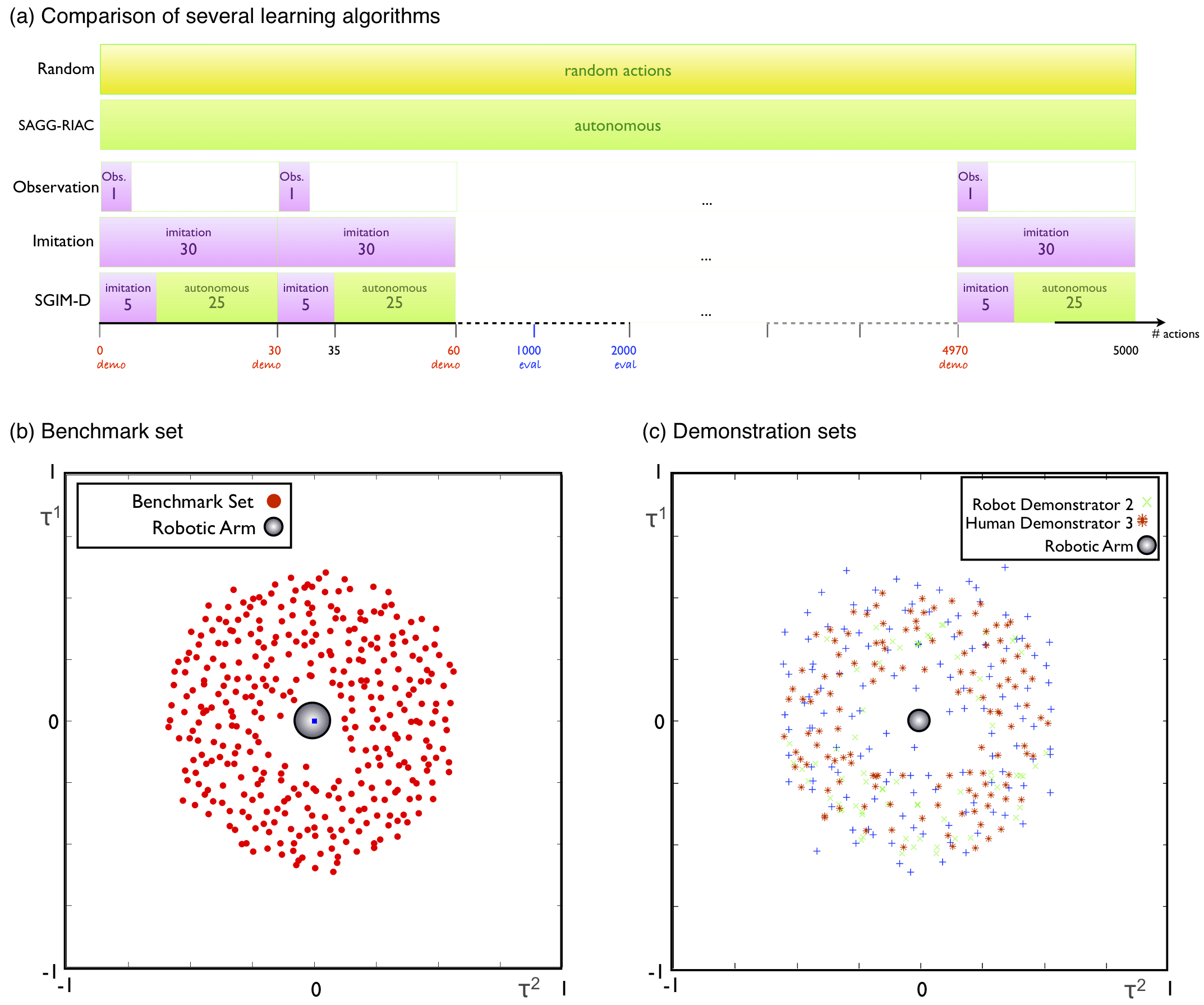}
\caption{(best seen in colors) (a): The experiment compares the performance of several exploration algorithms: Random exploration of the policy space A, autonomous exploration SAGG-RIAC, Learning from Observation, Imitation learning and SGIM-D. The comparison is made through the same experimental duration (5000 policies performed by the robot), through the same teaching frequency (every 30 policies) and through regular evaluation (every 1000 policies). \\
(b): Map in the 2D task space $T$ of the benchmark points used to assess the performance of the robot: by measuring how close they can reach each of these points. \\
(c): Maps in the 2D task space $T$ of the teaching sets used in SGIM-D, by three demonstrators. Demonstrator 1 is a SAGG-RIAC learner, while demonstrator 2 is an optimised SAGG-RIAC learner, and demonstrator 3 is a human teacher.}
\label{ExperimentalProtocol}
\end{figure*}

To assess the efficiency of SGIM-D, we decide to compare the performance of several exploration algorithms (fig. \ref{ExperimentalProtocol}a):
\begin{itemize}
\item Random exploration : throughout the experiment, the robot picks policies randomly in the policy parameter space $\Pi$.
\item SAGG-RIAC: throughout the experiment, the robot explores autonomously, without taking into account any demonstration by the teacher, and is  driven by intrinsic motivation   .
\item Imitation learning: every time the robot sees a new demonstration $\theta_{d}$ of the teacher, it repeats the policy while making small variations:   $\theta_{imitate} = \theta_{d} + \theta_{rand} $ with $\theta_{rand}$ a random movement parameter variation, so that $|\theta_{rand}|< \epsilon$. It keeps on repeating this demonstration until it sees a new demonstration every N policies, and then starts imitating the new demonstration.
\item Observation learning:  the robot does not make any policy, but only watches the teacher's demonstrations.
\item SGIM-D: the robot's behaviour is a mixture between Imitation learning and SAGG-RIAC. When the robot sees a new demonstration, it imitates the policy, but only for a short while. Then, it resumes its autonomous exploration, until it sees a new demonstration by the teacher. Its autonomous exploration phases take into account all its history from both the autonomous and imitation phases.
\end{itemize}

 For each experiment, we let the robot perform 5000 policies in total, and evaluate its performance every 1000 policies, using the method described below.

\subsection{Evaluation}

After several runs of Random explorations, SAGG-RIAC and SGIM-D, we determined the apparent reachable space basing on the set of all the reached points in the task/outcome space, which makes up some 300.000 points. We then tiled the reachable space into small rectangles, and generated a point randomly in each tile. We thus obtained a set of 358 goal points in the outcome space, representative of the reachable space (fig. \ref{ExperimentalProtocol}b). We will use these points to measure how close the system can get to each of these points with:

\begin{eqnarray}
mean_{\tau_g \in Benchmark Set}(D(\tau_g,\tau_{reached}))
\end{eqnarray}

 where $\tau_{reached}$ is the outcome observed by the robot when attempting to produce outcome $\tau_g$.

\subsection{Demonstrations}

For demonstrations, we used kinesthetics. The human teacher physically moves the robot, using both the physical robot and its model in the simulator. The model in the simulator is tele-operated by the teacher through the physical robot (\url{http://youtu.be/Ll_S-uO0kD0}). The human subject is presented with a grid of points to reach on the surface of the water, and he has to manipulate the physical robot to place the simulator's fishing rod nearest one of those point. After a habituation phase, we record the trajectories of each of the joints, and the position of the float when touching the surface of the water. We obtained a teaching set (fig. \ref{ExperimentalProtocol}c) from an expert teacher  of 127 samples. 

In order to analyse the specific properties of human demonstrations compared to random demonstrations in the SGIM-D algorithm, we also prepared two other sets of demonstrations, evenly distributed in the reachable space, and taken from a pool of data from several runs of SAGG-RIAC, using the previous SAGG-RIAC learners as teachers.

Thus we have 3 demonstration sets:
\begin{itemize}
\item demonstrator 1: SAGG-RIAC learners who now teach in return our SGIM-D robot. They choose demonstrations randomly among their memory exemplars $(\theta,\tau)$. It would illustrate the case of a naive teacher in a context of robot to robot teaching.
\item demonstrator 2:  SAGG-RIAC learners who now teach in return our SGIM-D, but carefully choose among their memory exemplars $(\theta,\tau)$ that are most reliable. The evenly distributed demonstrations minimise the variance of $\tau$ over several re-executions of the same policy  $\pi_\theta$. It would illustrate the case of a more evolute teacher in a scenario of robot to robot teaching. We built it taking inspiration from our observations of the demonstrator 3, to obtain a case halfway between the two other demonstrators in order to analyse the specific properties of human demonstrations.
\item demonstrator 3: a human teacher who tries to give demonstrations $(\zeta_{d}, \tau_{d})$ evenly distributed in the reachable space of $T$. These demonstrations are then processed by the learner  as explained in section \ref{CorrespondenceMapping}. The demonstrator was one of the authors, who however has no experience in fishing. The demonstrations used were captured only after a few attempt trials, therefore it does not give enough time to the demonstrator to get proficient at this fishing task.
\end{itemize}

Like with the evaluation set, we define a tile of the reachable space. The teacher observes the exploration of the learner, and gives to the learner a demonstration belonging to a subspace randomly chosen among those it has explored the least.
This teaching behaviour is a simple algorithm for active teaching, and can grow more elaborate taking inspiration from the field of Algorithmic Teaching \citep{Lopes2012A,Cakmak2010IICDL}.

The simulation data and analysis of the results are presented in the following section. 

\section{Results}

 For every simulation on the fishing experiment setup, 5000 movements are performed, and demonstrations taken from either of the 3 sets are given at fixed frequency every 30 movements. The performance was assessed on the same benchmark set every 1000 movements  (fig. \ref{ExperimentalProtocol}a).

\begin{figure*}
\centering
\includegraphics[width=\textwidth]{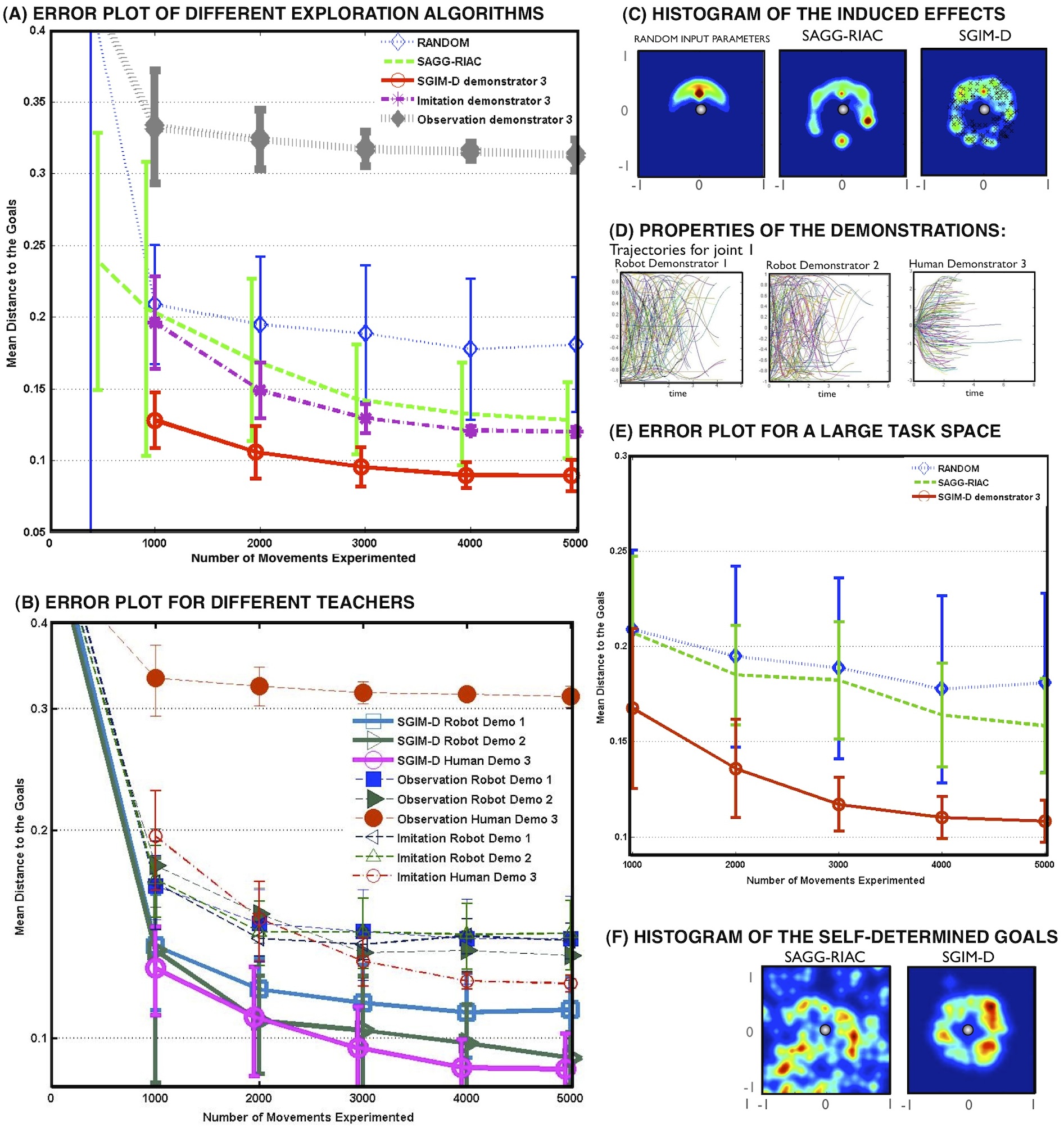}

\caption{ {\scriptsize
(best seen in colors)(a) : 
Evaluation of the performance of the robot under the learning algorithms: 
 random exploration, SAGG-RIAC, imitation and SGIM-D (for the human demonstrator 3. We plotted the mean distance to the benchmark points over several runs of the experiment with its variance errorbar.  \\
 (b): Evaluation of the performance of the robot learning with 3 different demonstrators, under the learning algorithms: 
 SGIM-D, Observation and Imitation. \\
 (c) : Histograms of the positions explored by the fishing rod inside the 2D task space $(\tau^1,\tau^2)$. Each column represents a different learning algorithm : random input parameters, SAGG RIAC and SGIM-D. We plotted the histogram for one example runs of the experiment. In the case of SGIM-D (3rd column), we also graphed the demonstrated outcomes with black crosses. \\
(d): Plot for the demonstrations of the trajectories for joint 1 (vertical axis: joint angles, horizontal axis: time).  \\
(e): Evaluation of the performance of the robot  in the case of a large task space  ($T= [-100,100]^2$ is $10^4$ times larger than the reachable space, but we only plotted here the distribution on the subspace $[-1,1]^2$), under the learning algorithms: 
 random exploration, SAGG-RIAC and SGIM-D. \\
(f): Distribution of all the goals set by the higher level during learning in a large space.  Each column shows the distribution of an experimental run of the SAGG-RIAC algorithm (col 1) or SGIM-D (col 2). 
  }
  }
\label{CompareEvaluationInterpolation}

\end{figure*}

\subsection{Better Precision}
Fig.  \ref{CompareEvaluationInterpolation}a represents how close the learner can get to any goal/task of the reachable space in $T$, at the same timestep of learning and, in the case of social learning, with the same amount of information given by the teacher. It plots the mean distance error of the attempts to reach the points in the benchmark set, with respect to the learning time (number of movements performed by the robot). The errors are averaged on all points in the benchmark, and also on different runs of the experiment. The 5  algorithms are ranked :
\begin{itemize}
\item Learning from Observation performs the worst: this is on the one hand due to the small number of samples, as the learner does not acquire experience on its own but only through observation of others. It is on the other hand due to the correspondence problems. Since the learner and teacher do not have the same policy primitives, the robot can not reproduce exactly the teacher's movements. 
\item RANDOM performs better because the learner acquires more data through its own experience, although the exploration is totally random. 
\item SAGG-RIAC decreases significantly the error value compared to Random Exploration. Not only has the asymptotic performance improved, but SAGG-RIAC also learns faster from the beginning.
\item Imitation Learning also decreases significantly the error value compared to Random Exploration. Its error level is comparable to SAGG-RIAC. Therefore, autonomous exploration, and learning that heavily depends on the teacher's demonstrations are comparable in terms of performance. We can note that the error variance of Imitation Learning is considerably smaller than that of SAGG-RIAC, because we use the same demonstrator with the same demonstration set, although the order of demonstrations changes. The error variance is likely to increase if we carry out our experiments with various demonstrators.
\item SGIM-D performs best and halves the error value compared to Random Exploration. Its asymptotic error approaches the noise level of the stochastic environment. Not only is the error level lower asymptotically, but it drops from the beginning of the learning process. SGIM-D performs better than pure autonomous exploration and pure socially guided exploration. 
\end{itemize}

The combination of autonomous exploration and socially guided exploration has thus bootstrapped the learning to decrease the performance error but also to improve the learning speed.

\subsection{A Wide Range of Outcomes}
To visualise the subspaces explored by each learning algorithm, we plot  the histogram of the positions of the float $\tau$ in the outcome space $T$ when it reaches the water (fig. \ref{CompareEvaluationInterpolation}c).  Each column represents a different algorithm, and we represented for each 2 example experiment runs. 
The 1st column  shows that a natural position lies around $\tau_c = (0, 0.5)$ in the case of an exploration with random movement parameters. Most movement parameters map to a position of the float around that central position. 
 The second column  shows the histogram in the task space of the explored points under SAGG-RIAC algorithm. Compared to random  exploration, SAGG-RIAC has increased the explored space, and most of all, covers more uniformly the explorable space. Besides, the exploration changes through time as the system finds new interesting subspaces to focus on and explore. Intrinsically motivated exploration has resulted in a wider repertoire for the robot. SGIM-D even emphasises this outcome: the explored space even increases further, with a broader range of  radius covered: the minimum and maximum distances to the centre have respectively decreased and increased.  Furthermore, the explored space is more uniformly explored, around multiple centres. 
 
The examination of the explored parts of $T$ show that random exploration only reaches a restricted subspace of $T$, while SAGG-RIAC and SGIM-D increase this explored space. 
This difference is mainly explained by the fact that most policies map to a restricted subspace of $T$, and on the contrary, the other parts of the reachable space can only be reached by a very small subset of policy parameters in $\Pi$. In other words, with random movements, the float has high chances of landing near that natural position. To make it reach other areas of the surface of the water, the arm needs to perform quite specific movements.
SGIM-D highlights these areas owing to its task space exploration and to  demonstrations. The teacher gives a demonstration that triggers the robot's interest and it is going to focus its attention on that area provided that local exploration improves its competence in this subspace. We also note that the demonstrations occurred only once every 30 movements. Even an occasional presence of the teacher, who does not need to monitor continuously the robot, can significantly improve the performance of the autonomous exploration.

\subsection{Dependence on Teachers}

Like any social learning method, SGIM-D's performance depends on the quality of the  demonstrations. Therefore, we need to examine further this dependency, and plot the mean error of the social learning algorithms for our 3 different demonstrators (fig.  \ref{CompareEvaluationInterpolation}b).   First of all, we notice that for all 3 teachers, SGIM-D performs better than the other algorithms. SGIM-D is therefore robust with respect to the quality of the demonstration as the teacher only guides the learner towards interesting policy or outcome subspaces, and the learner lessens its dependence on the teacher owing to self-exploration. Still, among the 3 demonstration sets we used, the demonstrations 1 that are chosen randomly bootstrap less than the demonstrations 2 that have smaller variance.  We also note that the human demonstrations (3), also bootstrap better than demonstrations 1, and slightly better than demonstrations 2. This result seems at first sight surprising, as the results of learning by observation seem to indicate the contrary: demonstrator 1 or 2 are more beneficial to the observation learner, since demonstrator 3's policies can be not easily reproduced due to correspondence problems.

To understand the reasons of this result, let us examine the different demonstrations. Fig. \ref{CompareEvaluationInterpolation}d plots the trajectories of the demonstrations.
We can see that demonstrations  show different distribution characteristics in the trajectory profile. The most noticeable difference is the case of demonstrator 3. Whereas the trajectories of demonstrators 1 and 2 seem disorganised, the joint value trajectories of demonstrator 3 are all monotonous, and seem to have the same shape, only scaled to match different final values.  

Indeed, the comparison of the demonstrations set 3 to random movements with ANOVA \citep{Krzanowski1988} indicates that we can reject the  hypothesis that demonstration set 3 comes from a random distribution ($p < 4.10^{-40}$). The demonstrations set 3 is not randomly generated but are well structured and regular.
Therefore, the human demonstrator shows a bias through his demonstrations to the robot, and orients the exploration towards different subspaces of the policy space. Indeed, the ANOVA analysis of the movements parameters $\theta$ performed during the learning reveals that they have different distributions with separate means. Because his demonstrations have the same shape, they belong to a smaller, denser and more structured subset of trajectories from which it is easier for the learner to generalise, and build upon further knowledge. Moreover, this comparative study highlights another advantage of SGIM-D: its robustness to the quality of demonstrated policies. The performance varies depending on the teacher, but still is significantly better than the SAGG-RIAC or imitation learner.

\subsection{Dependence on the Size of the Task Space}
To test whether our algorithms are scalable to large spaces, we plotted the mean distance error to the benchmark set, for a different task space (fig.  \ref{CompareEvaluationInterpolation}e).  This time,  the boundaries of each dimension have been multiplied by 100, which means that the size of $T$ has been multiplied by $10^4$. We can observe the effects on the performance of the SAGG-RIAC learner. Even though its mean error is lower than the random learner, it has increased compared to the case of the smaller task space. On the other hand, SGIM-D still learns to reach any point with good precision. Its mean error is significantly lower than the one of the SAGG-RIAC or the random learners.
Consequently, the social learning part of SGIM-D has helped it scale to larger spaces by allowing the robot to infer more quickly which parts of the task space are actually reachable and learnable.

\subsection{Identification of the Interesting Subspaces}
To investigate the reasons of  the difference in performance between SAGG-RIAC and SGIM-D, especially their different dependence on the task space size,  we can examine  the system's exploration of the task space. Fig. \ref{CompareEvaluationInterpolation}f plots the distribution of all the goals $\tau_g \in T$ chosen during the task space exploration of SAGG-RIAC and SGIM-D. The goals chosen by the SAGG-RIAC learner look disorganised, and cover all the task space $T$,  because it needs to sample at a minimum density before computing meaningful measures of interest, and find subspaces where it can actually learn. On the contrary, the SGIM-D learner only chooses its goals around the reachable space. Thus the teacher has helped the SGIM-D learner to identify and target the reachable space.

In conclusion, SGIM-D improves the precision of the system even with little intervention from the teacher, and helps point out key subregions to be explored. The teacher successfully transfers his knowledge to the learner and bootstraps autonomous exploration robustly, even in large task spaces. This bootstrapping is all the more efficient than the demonstrations chosen by the teacher enhance generalisation, for instance through similarity of the policies demonstrated. Although this paper has shown that SGIM-D can complete one type of goals only, studies \citep{Nguyen2012PPCRDC,Nguyen2012IICHR} have shown that it can learn in different kinds of task space.

The illustration experiments conducted showed good performance of SGIM-D in learning all the infinity of goals defined by the task space $T$, compared to pure autonomous exploration and social learning methods, in terms of precision and explored area. 
Moreover, analysis showed that on the one hand, it benefits from human teacher's demonstrations which orient its exploration towards small subspaces of policies and goals, and enable a faster identification of interesting subspaces. On the other hand, self experimentation helps it be more robust to demonstrations quality.

\section{Conclusion}

 This paper introduced Socially  Guided Intrinsic Motivation by Demonstration, \textbf{SGIM-D}, an architecture for online active learning of inverse models in continuous high-dimensional robotic sensorimotor spaces, and allowing a robot to learn multiple goals and generalise over a continuous ensemble of goals. SGIM-D efficiently combines social learning and intrinsic motivation strategies on both the policy and goal exploration levels. It actively samples goals while adapting to the difficulty of different subspaces. The analysis of the properties of this combination shows that the demonstrations structure orient the exploration towards a subspace of the policy space, independently of whether the demonstrations can be exactly reproduced by the learner or not. SGIM-D also takes advantage of the intrinsically motivated autonomous exploration to improve its performance and gain precision in the absence of the teacher for a wide range of outcomes/goals.  It is an original algorithm in that it is at the same time an active learning system of inverse models benefiting from human demonstrations, and also a PbD system which can learn and generalise to new goals. Our simulation indicates that SGIM-D successfully learns motor control even in an experimental setup as complex as having a  continuous 25-dimensional policy parameter space. We are now building the set-up to replicate these results with a physical robot and a real fishing-rod and to conduct a user study to assess how non-expert demonstrations influence the learning of our algorithm.

In this first step, for the sake of comparison of SGIM-D to other algorithms, we do not study further the effects of different parameters of  social interaction on the performance of the robot, for instance the impact of the frequency of the demonstrations given by the teacher. The parameters of the teaching, such as the rationales for selecting timing of the social interaction and demonstrations have not been chosen in this paper to optimise SGIM-D. A more precise study of these parameters has shown better performance of SGIM-D with different parameters \cite{Nguyen2012IICHR}. Moreover,  we could explore in depth the dependency of SGIM-D on the teacher, such as cases of sparse teachers, where the demonstrations belong to a small subspace only, or are in smaller number. Such studies would better illustrate the most general case when the human teacher can not perform everything, but is only proficient in a small subset of goals. We can also extend the work with a learner who self-determines whether to take into account a demonstration or not, taking inspiration from child psychology studies that show limitations of the role of parents\citep{Xu2011PII2JICDLICER}.

Most of all, we only considered a very simple interaction scenario between the learner and the teacher,  and we did not take into account interactive learning \citep{Chernova2009JAIR,Thomaz2006,Nicolescu2003}, where the learner asks for information when needed.
 More generally, exploring and evaluating systematically the other scenarios in which a human teacher can be involved, as mentioned in section III,  should be instructive. An interesting angle to study would also be the switching between mimicking, imitation and emulation behaviours. In this paper, the robot mimics the teacher for a fixed amount of time, and afterwards, SGIM-D takes into account these new data only from the goal point of view, as in emulation. However a more natural and autonomous algorithm for switching between or combining these different modes has been shown to improve the efficiency of the system in \cite{Nguyen2012PJBR}.

\begin{acknowledgements}
The authors would like to thank Paul Fudal, Jerome Bechu and Haylee Fogg for their support for the experimental setup, and Freek Stulp for his very helpful comments.
This research was partially funded by ERC Grant EXPLORERS 240007 and ANR MACSi.

\end{acknowledgements}

\bibliographystyle{spbasic}      
\bibliography{Nguyen_jrnl}

%
%

\end{document}